\documentclass{article}
\usepackage{iclr2027_conference,times}

\usepackage{hyperref}
\usepackage{url}
\usepackage{graphicx}
\usepackage{amssymb}
\usepackage{booktabs}
\usepackage{makecell}
\usepackage{caption}
\usepackage{tabularx}
\usepackage{multirow}
\usepackage{wrapfig}
\usepackage{amsmath}
\usepackage{array}
\usepackage{xcolor}
\usepackage{subcaption}
\usepackage{tikz}
\usepackage{soul}
\usepackage{longtable}
\usepackage{adjustbox}
\usepackage{enumitem}

\usepackage[most]{tcolorbox}
\tcbuselibrary{breakable}
\usepackage{listings}
\usepackage{soul, xcolor}

\newcommand{\sysname}{\textsc{DuplexAct}}

\newtcblisting{evalprompt}{
    listing only,
    breakable,
    colback=gray!6,
    colframe=gray!30,
    boxrule=0.4pt,
    arc=1.5pt,
    left=5pt,
    right=5pt,
    top=5pt,
    bottom=5pt,
    listing options={
        basicstyle=\ttfamily\footnotesize,
        breaklines=true,
        breakatwhitespace=true,
        breakindent=0pt,
        breakautoindent=false,
        columns=fullflexible,
        keepspaces=true,
        showstringspaces=false,
        aboveskip=0pt,
        belowskip=0pt
    }
}

\usepackage{etoolbox}

\makeatletter
\patchcmd{\@maketitle}
  {\begin{tabular}[t]{l}\bf\rule{\z@}{24pt}\@author\end{tabular}}
  {\centerline{%
    \begin{tabular}[t]{c}
      \rule{\z@}{24pt}\@author
    \end{tabular}%
  }}
  {}{}
\makeatother

\title{DuplexAct: How Interactive Are Frontier Full-Duplex Spoken Language Models?}

\author{
\textbf{Jiaheng Dong}$^{1,2}$,
\textbf{Shihong Tan}$^{3}$,
\textbf{Gongping Huang}$^{3}$,
\textbf{Jean Honorio}$^{1,4}$, \\
\textbf{Abhirup Ghosh}$^{2}$,
\textbf{Ting Dang}$^{1}$ \\[5pt]
$^{1}$The University of Melbourne
\qquad
$^{2}$University of Birmingham \\
$^{3}$Wuhan University
\qquad
$^{4}$ARC OPTIMA \\[5pt]
Correspondence: \texttt{jiahengd@student.unimelb.edu.au}
}

\iclrfinalcopy
\begin{document}

\maketitle
\lhead{Preprint}

\begin{abstract}
%\ag{Should we make the title a bit more slick? Like How interactive are the frontier Full duplex spoken language models?}

End-to-end (E2E) full-duplex spoken language models (FD-SLMs) enable simutanous listening and speaking continously, requiring models to take appropriate interaction actions as conversations unfold. However, existing benchmarks lack a structured action-level formulation, often evaluate action occurrence rather than success, and provide limited coverage of diverse real-world conversations. We introduce \textbf{\sysname{}}, a benchmark for evaluating interaction action taking in E2E FD-SLMs. \sysname{} defines a three-level action taxonomy spanning eight interaction actions and 24 corresponding interaction conditions. We further characterize successful action-taking through \emph{immediate execution} and \emph{subsequent follow-through}, and develop action-specific evaluation metrics for both. \sysname{} contains 4,691 test instances spanning casual conversation, meetings, customer service, and interviews, including 2,531 instances derived from real-world conversational recordings. Together, \sysname{} provides a structured and real-world-grounded framework for systematically assessing how E2E FD-SLMs participate in ongoing spoken interaction. Code and data are available at \url{https://github.com/JiahengDong/DuplexAct}.
\end{abstract}

\section{Introduction}
\vspace{-2pt}
End-to-end (E2E) full-duplex spoken language models (FD-SLMs) are advancing spoken human--AI interaction beyond rigid command--response exchanges toward more natural and collaborative communication \citep{wang2025ntpp, defossez2024moshi, yao2025flm}. By continuously listening while speaking, these models allow interaction to unfold without fixed boundaries between user and model turns, accommodating conversational patterns % characteristic of human conversation 
such as interruptions %, overlapping speech 
and backchannels.
%these models can remain engaged with incoming speech %even while producing their own responses, allowing interaction to unfold without fixed boundaries between user and model turns. This creates a fundamentally more dynamic setting, where models must accommodate patterns characteristic of human conversation, including interruptions, overlapping speech, and backchannels, as the conversation unfolds.

Navigating such interaction requires the model to %must 
continually determine whether, when, and how to participate in response to the evolving conversational context. Current E2E FD-SLMs typically learn %acquire 
interaction control implicitly from conversational sequence modelling, capturing %learning 
the temporal and contextual patterns that govern turn-taking from dialogue data. %participation from dialogue data such as when to take, hold, or yield a turn. 
Some models further introduce dedicated control tokens to represent specific decisions, %within generation, 
such as starting or stopping speech, maintaining silence, or producing a backchannel \citep{kim2026raon,cui2026minicpm,fang2026bayling}, embedding interaction control directly within the generative process.%  In this way, interaction control is learned and executed as part of the model's generative process.% rather than being governed by a separate dialogue controller.

However, %despite these advances in learning and implementing interaction control, 
\emph{current E2E FD-SLMs lack an explicitly defined and structured interaction-action space}, leading to %This will lead to 
fragmented action-level supervision and evaluation. %limiting both the systematic learning and comprehensive evaluation of models' interactive capabilities. 
Existing benchmarks~\citep{lin2025full-1, lin2026full-1.5,peng2025fdbench,he2026mtrduplexbench} further reflect this absence: they %therefore 
typically evaluate only a subset of actions, such as turn-taking, backchannels, and interruptions, or organize evaluation around specific phenomena such as overlapping speech, producing results that are neither comparable nor comprehensive at action level. % under different user intents. 
In contrast, human Interactional Competence (IC) research offers a more principled model: it structures conversational participation around clearly defined actions: turn management (starting, 
maintaining, and ending turns), interactive listening (backchannels and 
continuers), and breakdown repair~\citep{galaczi2018interactional}.  %around clearly defined interactional actions: starting, maintaining, and ending turns within turn management, providing backchannels and continuers within interactive listening, and repairing interactional breakdowns within breakdown repair \citep{galaczi2018interactional,may2020developing}.
Current benchmarks sample arbitrarily from this space, selecting a handful behaviors without principled categorization, limiting systematic understanding of both overall and action-specific model capabilities. 
%Consequently, the limited action coverage and mixed organization of actions, user intents, and interaction conditions in existing benchmarks limit systematic understanding of both overall model performance and action-specific capabilities.%\td{introduce figure 1?}

\begin{figure}[t]
    \centering
    \includegraphics[width=\linewidth]{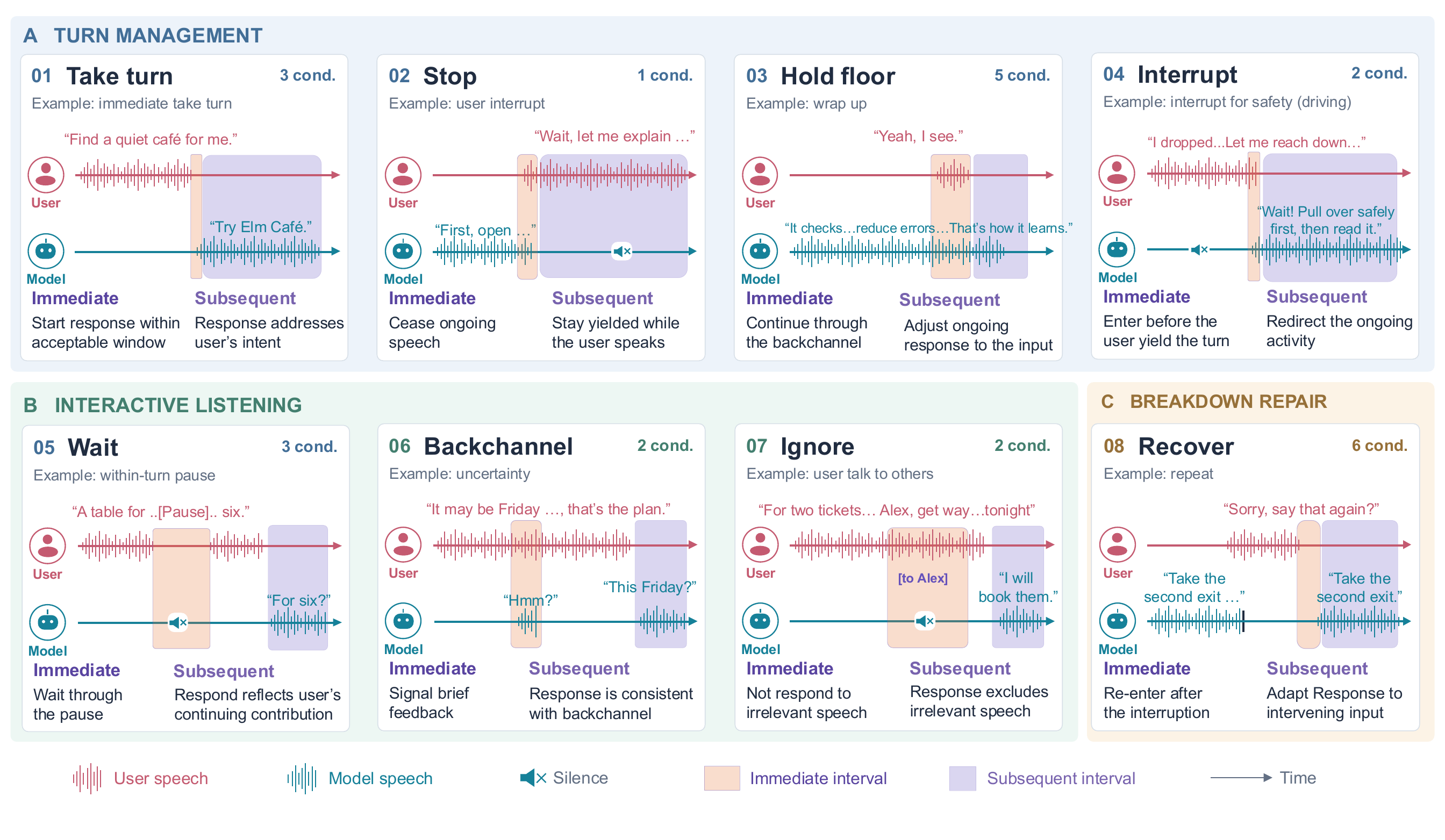}
    \vspace{-25pt}
    \caption{
    Overview of the \sysname{} interaction action space, with example conditions and high-level \emph{immediate execution} and \emph{subsequent follow-through} behaviours for each action.
    }
    \vspace{-15pt}
    \label{fig:duplexact_action_space}
\end{figure}

\emph{The absence of an structured action space further leaves the success criteria 
for actions under-specified.} Existing benchmarks rely on %commonly quantify observable interaction outcomes through 
metrics such as takeover rate, response latency, and backchannel frequency \citep{lin2025full-1}. These measures capture whether a behavior occurs, but not whether it succeeds. %While effective for measuring particular behaviours, these metrics provide limited characterization of whether an action is successfully executed as part of an evolving interaction. 
Both sequential decision-making and 
IC research make this point: an action is 
conditioned on the current conversational state and defined by its 
relation to both preceding context and subsequent 
response~\citep{deppermann_gubina_2026,may2020developing}. A backchannel, for instance, achieves successful interactive listening only if the listener's next turn develops the partner's ideas or asks a relevant follow-up~\citep{may2020developing}. 
% From a sequential decision-making perspective, an action is conditioned on the current conversational state and shapes what follows, where success cannot be read from a single moment in isolation. %affects the state and behaviour that follow. 
% Similarly, Interaction Competence (IC) and Conversation Analysis (CA) %  \td{spell out?}
% make the same point: actions are defined by their relation to both preceding context and subsequent response \citep{deppermann_gubina_2026,may2020developing}. Producing a backchannel alone does not establish successful interactive listening, which also requires developing the partner's ideas or asking relevant follow-up questions in the listener's next turn~\citep{may2020developing}. 
Evaluating interaction control therefore requires success criteria that capture two distinct dimensions: \emph{immediate execution}, whether the action is performed appropriately at the right moment, and \emph{subsequent follow-through}, whether the model's behavior remains consistent with that action as the interaction unfolds. Existing benchmarks miss the latter altogether and address the first only partially.
%Successful action taking should therefore account not only for whether the intended action is executed appropriately at the current moment, but also for whether subsequent model behaviour remains consistent with that action as the interaction develops. This motivates action-specific success criteria that capture both \emph{immediate execution} and \emph{subsequent follow-through}.

\emph{Datasets with complete action coverage and action-specific success annotations remain lacking. } 
%Beyond defining actions and their success criteria, reliable evaluation also requires sufficient coverage of the conditions under which each action should be performed.
Existing benchmarks fall short on three 
fronts: they cover only a subset of interaction actions, annotate occurrence rather than success, and rely heavily on synthetic or scripted interactions. These gaps compound: without a complete action space, annotations cannot capture whether each action is executed appropriately 
and followed through, only whether a surface behavior is present. Comprehensive evaluation therefore requires a dataset spanning the full action space with fine-grained success annotations across both execution 
and follow-through dimensions, grounded in real-world interactions and complemented by targeted synthesis where natural data are scarce.%\td{check}

% Existing benchmarks typically construct a limited set of conditions for individual interaction behaviors and concentrate on a narrow range of conversational settings, particularly casual or daily conversation. This coverage may obscure variation in action-taking performance across different conversational contexts, where interaction patterns %and expectations can 
% differ. They also %Moreover, existing evaluations r
% rely substantially on synthetic, scripted, or simulated interactions, leaving naturally occurring data underrepresented. %while naturally occurring conversational data remain comparatively underrepresented. 
% Comprehensive action-level evaluation therefore requires broad coverage across action-specific conditions and conversational scenarios, grounded where possible in real-world interactions and complemented by targeted synthesis for conditions that are difficult to obtain naturally.

To address these limitations, we propose \textbf{\sysname{}}, benchmarking interactive capabilities of E2E FD-SLMs. Grounded in human Interactional Competence,  %IC and CA, 
\sysname{} provides a structured formulation of interaction actions in Figure~\ref{fig:duplexact_action_space}, with action-specific definitions and evaluation criteria, diverse conversational scenarios, 
and real-world spoken interactions. %together with broad coverage of diverse conversational scenarios and real-world spoken interactions. 
%Our results reveal highly uneven capabilities across actions and conditions: no model performs well across the full action space, and targeted supervision produces only selective strengths.  Crucially, successful immediate execution frequently fails to carry through to the subsequent behavior required by an action, exposing the gap between performing an action and sustaining its interactional consequences.
Our contributions are: 
\vspace{-1pt}
\begin{itemize}[leftmargin=1.5em, labelsep=0.4em]
    \vspace{-1pt}
    \item \textbf{A structured interaction-action taxonomy.} We propose a structured 3-level hierarchical taxonomy spanning eight interaction actions across 24 conditions that provide a principled framework %an action-level framework 
    for evaluating how full-duplex SLMs participate in spoken interaction.
    \vspace{-1pt}
    \item \textbf{Action-specific success criteria.} We introduce a two-part structure characterising successful action-taking through 
    \emph{immediate execution} and \emph{subsequent follow-through}, with corresponding metrics for each action. %that characterizes successful action taking through \emph{immediate execution} and \emph{subsequent follow-through}. %We define the requirements of each interaction action grounded in IC framework and derive corresponding action-specific evaluation metrics for both components.
     \vspace{-1pt}
    \item \textbf{A large-scale, real-world-grounded evaluation dataset.} We provide a dataset with action-level annotations, consisting of 4,691 instances spanning 24 interaction conditions across four conversational scenarios, with 54\% real-world interactions. % casual conversation, meetings, customer service, and interviews. %The dataset grounds 54\% of its instances in real-world conversational recordings, complemented by targeted synthesis to ensure coverage of underrepresented action--condition combinations.
     \vspace{-1pt}
     \item \textbf{A comprehensive evaluation powered by the action design.} Evaluating 14 models, we reveal uneven performance across the action space, weak subsequent reasoning beyond immediate action-taking, and limited adaptation to conversational norms across scenarios.
    % \item \textbf{A comprehensive evaluation of current E2E FD-SLMs} We evaluate 14 models and reveal limitations in current full-duplex training, including incomplete coverage of the action space, weak alignment between action taking and subsequent reasoning, and limited adaptation to conversational norms across scenarios.

\end{itemize}
\vspace{-10pt}
\section{Related Work}

\paragraph{End-to-End Full-Duplex SLMs. }
Recent E2E FD-SLMs internalize interaction control within the generative model, learning turn-taking and other conversational dynamics from %time-synchronous 
dialogue data \citep{defossez2024moshi,fang2026bayling}. These decisions are often represented through model-specific control tokens or signals: %Moshi uses \texttt{PAD}/\texttt{EPAD} for speech initiation \citep{defossez2024moshi},
BayLing-Duplex uses \texttt{[SILENCE]}/\texttt{[ASSISTANT]} for dialogue-state control \citep{fang2026bayling}, while Raon-SpeechChat further introduces an \texttt{audio\_output\_backchannel} token to distinguish backchanneling from general speech initiation~\citep{kim2026raon}. %\td{ref}. 
Lychee-FD similarly uses a dedicated control head with \texttt{Start}/\texttt{Stop} signals~\citep{liu2026lycheefd}. %\td{ref}. 
However, these controls are defined at different granularities and for individual model requirements, lacking a %explicit and 
structured and common interaction-action space for systematically representing and evaluating conversational actions.

\begin{table*}[t]
\centering
\caption{
Comparison of existing benchmarks.
Action coverage: $\checkmark$ explicitly covered;
$\circ$ partial coverage,
and $\times$ no coverage.
Scenarios: C = casual conversation, S = customer service,
I = interview, M = meeting; -- = unclassified.
$^{*}$Multi-turn conversations.
}
%\vspace{-5pt}
\label{tab:benchmark_comparison}

\begingroup
\scriptsize
\setlength{\tabcolsep}{2.3pt}
\renewcommand{\arraystretch}{1.0}

\begin{adjustbox}{max width=\linewidth}
\begin{tabular}{@{}lccccccccccccc@{}}
\toprule
\textbf{Benchmark}
& \multicolumn{2}{c}{\textbf{Evaluation}}
& \multicolumn{3}{c}{\textbf{Data}}
& \multicolumn{8}{c}{\textbf{Action Coverage}}
\\
\cmidrule(lr){2-3}
\cmidrule(lr){4-6}
\cmidrule(lr){7-14}

& \textbf{Immediate}
& \textbf{Subsequent}
& \textbf{Real}
& \textbf{Total}
& \textbf{Scenario}
& \textbf{Wait}
& \textbf{Ignore}
& \textbf{Take Turn}
& \textbf{Backch.}
& \textbf{Stop}
& \textbf{Hold Floor}
& \textbf{Recover}
& \textbf{Interrupt}
\\
\midrule

FDB-v1
& $\checkmark$
& $\times$
& 54\%
& 727
& C
& $\circ$
& $\times$
& $\checkmark$
& $\checkmark$
& $\circ$
& $\times$
& $\circ$
& $\times$
\\

FDB-v1.5
& $\checkmark$
& $\times$
& 0\%
& 499
& C
& $\times$
& $\times$
& $\times$
& $\times$
& $\circ$
& $\circ$
& $\circ$
& $\times$
\\

FDB-v2.0
& $\checkmark$
& $\times$
& 0\%
& 200$^{*}$
& --
& $\times$
& $\times$
& $\circ$
& $\circ$
& $\circ$
& $\circ$
& $\circ$
& $\times$
\\

FDB-v3.0
& $\checkmark$
& $\times$
& 0\%
& 100$^{*}$
& S
& $\times$
& $\times$
& $\circ$
& $\times$
& $\times$
& $\times$
& $\times$
& $\circ$
\\

HumDial-FDBench
& $\checkmark$
& $\times$
& 0\%
& 5,000
& --
& $\circ$
& $\circ$
& $\times$
& $\times$
& $\circ$
& $\circ$
& $\circ$
& $\times$
\\

$\tau$-Voice
& $\checkmark$
& $\times$
& 0\%
& 278
& S
& $\times$
& $\times$
& $\circ$
& $\times$
& $\circ$
& $\circ$
& $\circ$
& $\circ$
\\

FLEXI
& $\checkmark$
& $\times$
& 0\%
& 1,200
& --
& $\circ$
& $\times$
& $\checkmark$
& $\checkmark$
& $\checkmark$
& $\circ$
& $\circ$
& $\checkmark$
\\

FD-Bench
& $\checkmark$
& $\times$
& 0\%
& 293$^{*}$
& C
& $\times$
& $\times$
& $\circ$
& $\times$
& $\circ$
& $\circ$
& $\circ$
& $\circ$
\\

MTR-DuplexBench
& $\checkmark$
& $\times$
& 33\%
& 600$^{*}$
& C
& $\circ$
& $\times$
& $\checkmark$
& $\checkmark$
& $\circ$
& $\times$
& $\circ$
& $\times$
\\

\midrule

\textbf{\sysname{}}
& $\checkmark$
& $\checkmark$
& \textbf{54\%}
& \textbf{4,691}
& \textbf{C,S,I,M}
& $\checkmark$
& $\checkmark$
& $\checkmark$
& $\checkmark$
& $\checkmark$
& $\checkmark$
& $\checkmark$
& $\checkmark$
\\

\bottomrule
\end{tabular}
\end{adjustbox}

\endgroup
\end{table*}

\vspace{-6pt}
\paragraph{Full-Duplex SLM Benchmarks}
Existing full-duplex benchmarks, compared in Table~\ref{tab:benchmark_comparison}, lack a structured action space and evaluate only arbitrary subsets of conversational actions. Moreover, they mix model actions and the conversational conditions under which those actions are expected to occur, at a single flat level. %mixing model actions, user intents, and conversational conditions as parallel evaluation dimensions. 
FDB v1 places model actions such as turn-taking and backchanneling alongside conversation conditions such as user pauses and user interruptions \citep{lin2025full-1}. FDB~v1.5 and FD-Bench organize evaluation around overlapping-speech conditions such as user interruption and background speech~\citep{lin2026full-1.5,peng2025fdbench}, while HumDial-FDBench evaluates rejection across conditions including third-party speech~\citep{wang2026humdialfdbench}. %\td{check if this is what you want to convey?} 
Mixing \emph{what action the model should perform} with \emph{the conditions under which it applies}, which operate at different levels of granularity, limits systematic action-level characterization of model capabilities. A structured, multi-level action space is therefore required.

\vspace{-6pt}
\paragraph{Evaluation Metrics. }%Existing evaluation metrics similarly focus on local, observable interaction\td{jargon, what are they?} outcomes rather than action-specific definitions of success. Common measures include takeover or yield rates, response and stop latency, backchannel occurrence, and response quality \citep{lin2025full-1,lin2026full-1.5,peng2025fdbench}. These metrics effectively capture whether an expected behaviour occurs and how promptly it is executed, but provide limited information about whether the intended action is successfully carried through as the interaction develops.
Existing metrics measure whether a behavior occurs and how promptly such as
takeover rates, response latency, and backchannel 
occurrence~\citep{lin2025full-1,lin2026full-1.5,peng2025fdbench}, but not whether the action is successfully carried through 
as the interaction unfolds. A model that backchannels disruptively and one that backchannels 
appropriately score identically on occurrence, but only one is listening. Evaluations should therefore assess not just whether an action occurs, but whether it succeeds, e.g., performed appropriately and followed through (Table ~\ref{tab:benchmark_comparison}).

\section{Constructing \sysname{}}
\vspace{-5pt}

\begin{figure*}[t]
    \centering
    \includegraphics[width=\textwidth]{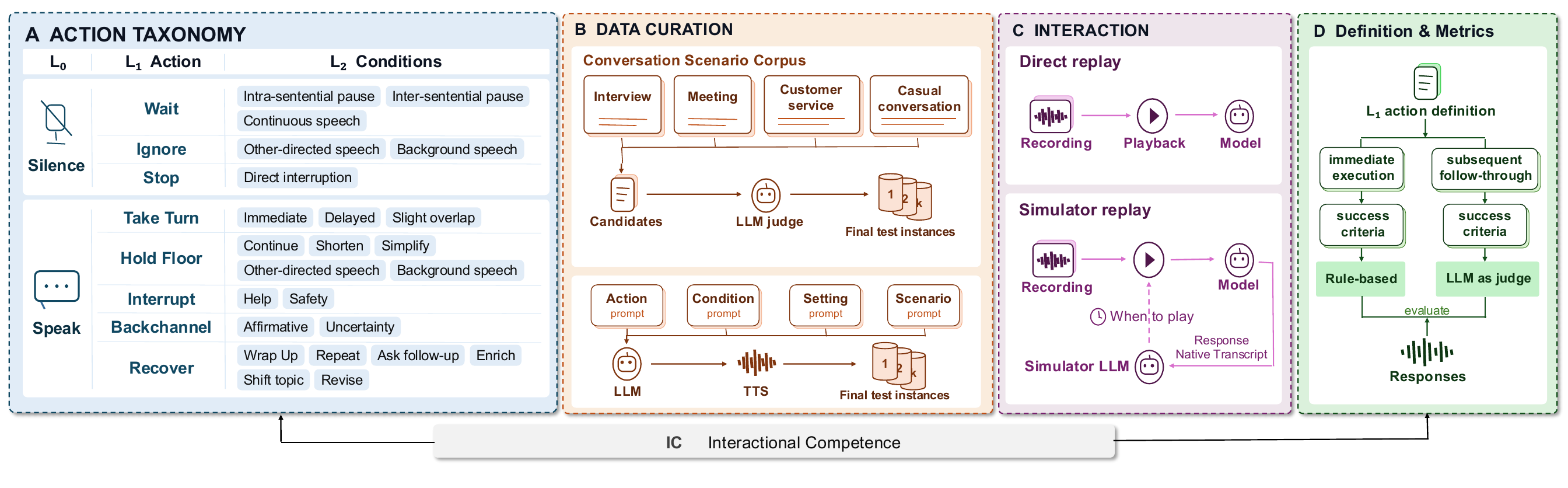}
    \vspace{-15pt}
    \caption{
    Overview of the \sysname{} benchmark framework.
    %\textbf{(A)} The action taxonomy consists of 8 actions and 24 conditions.
    %\textbf{(B)} Test instances are curated from real-world conversational corpora and synthesized data.
    %\textbf{(C)} Direct Replay and Simulator Replay interaction pipeline. \td{interaction evaluation?}
    %\textbf{(D)} Model responses are evaluated through immediate execution and subsequent behavior. %\td{d should reflect your immediate and subsequent-follow up point?}
    }
    \vspace{-8pt}
    \label{fig:framework}
\end{figure*}

We construct \sysname{} to investigate the interactive capabilities of FD-SLMs through a structured interaction-action taxonomy and action-specific success criteria, supported by a purpose-built dataset 
grounded in real-world interactions. %their ability to take appropriate interaction actions in diverse conversational settings, with 
The overall framework is illustrated in Figure~\ref{fig:framework}.

%\subsection{Benchmark Construction}
\vspace{-5pt}
\subsection{Interaction Action Space and Condition Construction}
\label{sec:action_space}
\vspace{-5pt}

We formulate full-duplex interaction as a three-level taxonomy separating participation state (L0), interaction actions (L1), and the conditions (L2) under which actions are required (Figure~\ref{fig:framework}). \textbf{L0} represents the model's observable participation state: \emph{speak} or \emph{silence}. While L0 is directly observable, it alone is insufficient for evaluation.

\textbf{L1} defines eight interaction actions that form the core of our evaluation framework. We consolidate and adapt these from three categories in human interaction research: turn management, interactive listening, and 
breakdown repair~\citep{galaczi2018interactional,may2020developing}. Turn management covers starting, maintaining, ending, and interrupting a turn, 
adapted as \emph{Take Turn}, \emph{Hold Floor}, \emph{Stop}, and \emph{Interrupt}. Interactive listening, concerned with how a listener displays recipiency while another participant speaks, is realized as \emph{Backchannel} for active listenership, and \emph{Wait} and \emph{Ignore} for basic and selective listening in continuous human--model interaction. Breakdown repair, which encompasses responses to interactional 
disruption, is abstracted as \emph{Recover}, capturing how the model should behave after being interrupted. 

\textbf{L2} specifies the conversational conditions under which each L1 action is required, yielding 24 action--condition pairs. This separation enables evaluation of not only whether a model can perform an action, but whether it can recognize the condition and adapt its action accordingly: \emph{Take Turn} after a completed but complex or simple turn, and an explicit invitation each require different turn-entry behaviour from the model. %This separation enables evaluation of not just whether a model performs an action, but whether it recognizes when to do so: \emph{Take Turn} after a completed turn, a mid-utterance pause, and an explicit invitation each require different behavior from the model.  
We systematically consolidate conditions from prior benchmarks and model 
evaluations~\citep{lin2025full-1,lin2026full-1.5,peng2025fdbench,ge2025flexi,wang2026humdialfdbench,chiang-etal-2026-shanks} and extend 
them to underrepresented settings such as safety-related \emph{Interrupt}, for instance, spans both home and driving domains. Organizing these conditions within a unified taxonomy makes action--condition coverage explicit and comparable across models for the first time. Detailed mappings are provided in Appendix~\ref{app:action_space}.%\jd{adjusted the take turn here}%\td{double check, especially for L2.}

\subsection{Benchmark Data Curation}
\label{sec:data}
\vspace{-3pt}
We construct dataset to cover the 24 action--condition pairs defined in our taxonomy, drawing primarily from real-world conversational corpora and resorting to controlled synthesis only when natural data are insufficient for a particular condition. Instances span four conversational 
scenarios: casual conversation, meetings, customer service, and interviews, representing distinct interaction norms and 
expectations~\citep{schegloff2007sequence,femo2026short}, as shown in Figure~\ref{fig:framework}B.% each providing a different context in which the actions are elicited and evaluated.

% We primarily construct test instances from real-world conversational corpora, using controlled synthesis when data are insufficient for particular L2 conditions. We cover four conversational scenarios, including casual conversation, meetings, customer service, and interviews, representing different interaction norms and expectations~\cite{schegloff2007sequence,femo2026short}. 

\vspace{-6pt}
\paragraph{Real-world interaction extraction.}
We collect naturally occurring conversations from CANDOR~\citep{reece2023candor} for casual conversation, AMI~\citep{kraaij2005ami} for meetings, Loqui~\citep{passonneau2014loqui} for customer service, and CORAAL~\citep{kendall2023coraal} for interviews. For each L2 condition, candidate segments expressing the target action are first identified 
through action-specific signals: speaker transitions for turn-taking actions, overlapping speech for interruption and backchannel, and user speech without model response for listening actions. Each candidate is then verified by GPT 5.4, which confirms the intended action is occurring 
and assigns the correct L2 condition label. Only segments passing both stages are retained. %First, rule-based structural criteria ($C_1$)\td{what does this mean? }retrieve candidate segments for each L1 action using dataset timestamps, transcripts, and available human annotations of conversational events such as backchannels. Second, an LLM-based semantic criterion ($C_2$) verifies the intended action, assigns the corresponding L2 condition label, and checks whether the extracted interaction is conversationally complete. \td{do we need to introduce c1 and c2 here? just describing the process?}\td{Candidate segments expressing the target action are first identified 
%through action-specific signals: speaker transitions for turn-taking actions, overlapping speech for interruption and backchannel, and user speech without model response for listening actions. Each candidate is then verified by an LLM, which confirms the intended action is occurring and assigns the correct L2 condition label. Only segments passing both stages are retained.}

\vspace{-6pt}
\paragraph{Controlled interaction synthesis.}
When existing conversational corpora do not provide sufficient samples for an L2 condition, we generate controlled conversational transcripts using GPT-5.4, following the transcript-to-speech synthesis pipeline adopted in $\tau$-Voice~\citep{ray2026tauvoice}. As illustrated in Figure~\ref{fig:framework}, generation is controlled by four prompts specifying the \emph{scenario}, \emph{action}, \emph{condition}, and \emph{setting}, where the setting prompt varies conversation topics and physical environment to increase diversity within each scenario. The generated transcripts are then rendered into speech using ElevenLabs v3 %\td{ref}
with diverse realistic personas following the configuration in $\tau$-Voice. Detailed generation prompts and synthesis configurations are provided in Appendix~\ref{app:controlled_synthesis} and~\ref{app:model_setup}.%\td{details of the prompt in ?}

\begin{wraptable}{r}{0.34\textwidth}
\vspace{-8pt}
\centering
\small
\caption{Benchmark statistics.}
\vspace{-8pt}
\label{tab:benchmark_statistics}
\begin{tabular}{lr}
\toprule
\textbf{Statistics} & \textbf{Number} \\
\midrule
Actions & 8 \\
Conditions & 24 \\
Real-world instances & 2,531 \\
Synthesized instances & 2,160 \\
\midrule
Total instances & 4,691 \\
Total audio duration & 12.44 h \\
\bottomrule
\end{tabular}
\vspace{-6pt}
\end{wraptable}

\paragraph{Human Validation.} We further conduct human validation with 20 annotators on 10\% of the test instances. Majority-vote agreement reaches 97.9\% for L1 action labels and 95.5\% for L2 condition labels, confirming the reliability of the extraction and synthesis pipelines. %By majority vote, raters validated 97.9\% of L1 action labels and 95.5\% of L2 condition labels. 
%For synthesized instances, conversational content and speech achieve average naturalness ratings of 3.96 and 3.43 out of 5\td{not natural, hard to explain, put in appendix, like detailed human evaluations in xx?}, with non-negative rates\td{what is this?} of 93.1\% and 76.8\%. 
Table~\ref{tab:benchmark_statistics} summarizes the overall benchmark statistics, including 4,691 instances over 12.44 hours. Detailed validation and data are provided in Appendices~\ref{app:human-validation} and~\ref{app:data_curation}.

\subsection{Evaluation}
%\ag{To me the first question for evaluation would be: given that the conversation data contains speech from two humans, how do you evaluate a model performance? Can we say this in a sentence or two here?}
% Our evaluation converts human--human interactions into model test instances by treating one participant's speech as the model input and using the observed interaction of the model to test whether its behaviour satisfies the corresponding action-specific requirements.

\subsubsection{Action Success Definition and Metrics}
\label{sec:action_evaluation}
\vspace{-3pt}

Successful action-taking is evaluated at two stages as shown in Figure~\ref{fig:framework}D. \emph{Immediate execution} measures whether the expected action occurs at the target interaction point and with appropriate timing. %, consistent with how prior work evaluates interaction behaviour.\td{is this true?} 
 We further propose \emph{Subsequent follow-through}, measuring whether the ensuing interaction remains consistent with that action as the conversation develops. This two-stage formulation reflects the sequential organization of human interaction, where the appropriateness of an action is validated not 
by its occurrence but by the conduct that follows~\citep{galaczi2018interactional,may2020developing, 
deppermann_gubina_2026}.%\td{check; while you write the method, you also want to highlight what is novel.}

% \vspace{-6pt}
% \paragraph{Action-specific evaluation.}
We evaluate each test instance against action-specific success criteria at two stages, as summarized in Table~\ref{tab:action_evaluation}. \emph{Immediate execution} is determined automatically 
by rule-based criteria, checking whether the model produces the expected action at the correct interaction point. For example, whether it takes the turn within the expected response window or falls silent upon a stop signal. \emph{Subsequent follow-through} is assessed by GPT-5.4 against an action-specific checklist, checking whether the model's behavior after the action remains consistent with it. For example, whether a model that recovers after an interruption adapts its resumed response accordingly, or whether one that backchannels uncertainty asks a relevant follow-up question.
%took the turn maintains a coherent and relevant response, or whether one that backchannelled asks a relevant follow-up.

We report two success rates accordingly. \emph{Immediate execution success rate} ($I$) is the proportion of instances satisfiy the immediate criterion, while \emph{subsequent follow-through success rate} ($S$) is the proportion of immediate-success instances that also satisfy the corresponding subsequent criterion. Both rates are micro-averaged across the L2 conditions associated with each action, with condition-specific constraints applied where required. A model may achieve high $I$ but low $S$ when it performs the expected action at the interaction point but fails to sustain the behaviour that should follow, suggesting the model may learn the surface pattern of an action without reasoning how that action should embed in and shape the conversational flow that follows. This is precisely the gap that metrics based only on immediate occurrence cannot reveal. Detailed criteria and metrics are in Appendix~\ref{app:action_evaluation}%exposing a gap that metrics based only on immediate action occurrence cannot reveal. 

%We report these as two rates. Let $N(a,c)$ denote the number of instances for action $a$ under condition $c$. The immediate execution rate $I_a$ and subsequent follow-through rate $S_a$ are:\td{include equation number using equation}
%\[
%I_a =
%\frac{\sum_{c\in C_a}\operatorname{succ}_I(a,c)}
%{\sum_{c\in C_a}N(a,c)},
%\qquad
%S_a =
%\frac{\sum_{c\in C_a}\operatorname{succ}_S(a,c)}
%{\sum_{c\in C_a}\operatorname{succ}_I(a,c)}.
%\]
%\td{are these equations really meaningful? without defining these succ terms, it may not make much sense and also not highlight any of the novelty? }Here $S_a$ is conditioned on successful immediate execution. A 
%model that executes actions correctly but fails to sustain them will 
%score high on $I_a$ but low on $S_a$, exposing a gap that 
%occurrence-based metrics cannot reveal. 
%For each test instance, immediate success is determined automatically by wheter the model passes the rule-based immediate success criterion, whereas subsequent success is determined using a single-item, action-specific checklist with GPT-5.4 as judge. 

%We additionally report timing diagnostics where applicable: yield latency for \emph{Stop}, interruption timing difference for \emph{Interrupt}, and recovery latency for \emph{Recover}.\td{does this belong to immediate evalautoons? why mentioning these here? } Detailed timing definitions and judging criteria are provided in Appendix~\ref{app:action_evaluation}.

\begin{table*}[t]
\centering
\caption{
Action-specific success criteria and condition-specific constraints in \sysname{}.
}
\vspace{-5pt}
\label{tab:action_evaluation}

\scriptsize
\setlength{\tabcolsep}{4pt}
\renewcommand{\arraystretch}{1.12}

\begin{tabular}{
p{0.09\textwidth}
p{0.09\textwidth}
p{0.56\textwidth}
p{0.18\textwidth}
}
\toprule
\textbf{Action}
&
\textbf{Stage}
&
\textbf{Success Criterion}
&
\textbf{Condition Constraint}
\\
\midrule

\textbf{Wait}
&
Immediate
&
Remain silent while the user retains the floor.
&
Wait interval\\
\cmidrule(lr){2-4}
&
Subsequent
&
Respond in a way that reflects the continual conversation context. %user's continuing contribution.
&
Wait interval\\
\midrule

\textbf{Ignore}
&
Immediate
&
%Remain silent to speech that does not require model participation.
Remain silent to non-relevant speech while the user holds the floor.
&
None\\
\cmidrule(lr){2-4}
&
Subsequent
&
Respond 
% to the actual user intent 
% without incorporating 
leaving the information from ignored part of the conversation.
%Respond to the addressed user content without incorporating the ignored speech.
&
None\\
\midrule

\textbf{Take Turn}
&
Immediate
&
Begin a substantive response.
&
Response interval
\\
\cmidrule(lr){2-4}
&
Subsequent
&
Address user's intent while respecting the information, constraints, and revisions 
% in the user speech.
&
None\\
\midrule

\textbf{Backchannel}
&
Immediate
&
Produce a brief feedback signal while allowing the user to retain the floor.
&
Backchannel function\\
\cmidrule(lr){2-4}
&
Subsequent
&
Maintain subsequent behaviour consistent with the expressed feedback.
&
None\\
\midrule

\textbf{Stop}
&
Immediate
&
Terminate ongoing speech after the user interrupts.
%initiates an interruption.
&
None\\
\cmidrule(lr){2-4}
&
Subsequent
&
Remain silent %yielded 
while the user's intervention continues.
&
None\\
\midrule

\textbf{Hold Floor}
&
Immediate
&
%Maintain the ongoing contribution without yielding the floor.
Continue speaking without yielding the floor.
&
None\\
\cmidrule(lr){2-4}
&
Subsequent
&
Continue or adapt the ongoing response based on concurrent speech intent.
&
Response adaptation type\\
\midrule

\textbf{Recover}
&
Immediate
&
Re-enter the conversation after pausing for an interruption.%yielding to an interruption.
&
None\\
\cmidrule(lr){2-4}
&
Subsequent
&
Adapt the resumed response %contribution 
following the interruption.
&
Response adaptation type\\
\midrule

\textbf{Interrupt}
&
Immediate
&
Take the floor while the user is speaking.&
None\\
\cmidrule(lr){2-4}
&
Subsequent
&
Provide interruption content appropriate based on the user's intent or situation. %Carry out the required intervention.
&
Intervention purpose
\\

\bottomrule
\end{tabular}
\vspace{-7pt}
\end{table*}

\subsubsection{Interaction Pipeline}
\label{sec:interaction_pipeline}
\vspace{-3pt}

% We use two interaction pipelines depending on how each action is triggered, preserves natural interaction dynamics while maintaining a controlled and reproducible benchmark. 
We use two interaction pipelines depending on how each action is triggered: Direct Replay, where the model responds to a fixed audio prompt, and Simulator Replay, where a user simulator drives live bidirectional interaction. Both preserve natural interaction dynamics while maintaining a controlled and reproducible evaluation setting.

%For actions with predefined timing, such as \emph{Wait} and \emph{Ignore}, where the user speaks and the model listens, user audio is replayed directly to the model. For actions whose timing depends on the model's ongoing behavior, such as \emph{Stop} and \emph{Interrupt}, a simulator generates user input dynamically in response to the model's output. This design preserves natural interaction dynamics while maintaining controlled and reproducible benchmark conditions. 

\vspace{-6pt}
\paragraph{Direct replay.}
As shown in Figure~\ref{fig:framework}C, for \emph{wait}, \emph{ignore}, \emph{take turn}, and \emph{backchannel}, the evaluation condition is fully determined by the input audio and does not rely on the model's ongoing response. We therefore use direct replay, where each instance is streamed to the model at its required input frame rate while model output is recorded concurrently. 

\vspace{-6pt}
\paragraph{Simulator-based interaction.}
For \emph{Stop}, \emph{Hold Floor}, \emph{Recover}, and \emph{Interrupt}, the user-side interaction timing must adapt to the model's ongoing response and therefore cannot be predetermined by a fixed input timeline. %For \emph{stop}, \emph{hold floor}, \emph{recover}, and \emph{interrupt}, evaluation requires interaction events whose timing depends on the model's ongoing behaviour. \ag{Why? It seems in these cases the speaker needs to adapt how the agent under test replies. This can not be controlled in real datasets. If yes, then say that.} 
We use an LLM-based simulator to decide \emph{when} to inject a prerecorded user cue, while the cue content 
and audio remain fixed by the benchmark instance. The simulator observes the model's streaming transcript together with the reference response and cue information to make this timing decision, avoiding online speech synthesis and reducing latency. 
%We therefore use an LLM-based interaction simulator to determine when the prerecorded cue should be injected. Rather than generating user responses and synthesizing speech online, we restrict the LLM simulator to deciding \emph{when} a prerecorded cue should be injected, while the cue content and audio remain fixed by the benchmark instance. This design improves interaction controllability and reduces simulator-induced latency by avoiding online response generation and speech synthesis. To make this timing decision, the simulator observes the model-provided streaming transcript together with the reference response and cue information from the benchmark instance. We use this transcript as the simulator signal rather than retranscribing the rendered audio, aligning cue insertion with the model's ongoing generation while reducing timing variation introduced by downstream buffering and playback. %\ag{Right, this description is unclear. One problem I think is that it feels you are testing the agent under test using another AI based on their judgment of timing and content. Can we have a figure to illustrate this?} 
This separates \emph{when} the intervention occurs from \emph{what} is said, preserving controlled benchmark content while adapting interaction timing to each model's live generation. Detailed streaming, synchronization, cue-insertion, and simulator configurations are provided in Appendix~\ref{app:interaction_pipeline}.

\section{Experimental Setup}
\label{sec:experimental_setup}
\vspace{-5pt}
\paragraph{Evaluated systems.} %\ag{citations? Why do we choose them? Do we follow other benchmarks?} \jd{Added citations and reasons of why these are chosen, but for new released models, there is no survey or leader board has includes them.}

We evaluate 14 recent FD-SLMs: 10 open-source E2E models and four closed-source models.
The open-source models comprise \emph{Moshi}~\citep{defossez2024moshi}, \emph{Freeze-Omni}~\citep{wang2025freezeomni}, \emph{VITA-1.5}~\citep{fu2026vita}, and \emph{Mini-Omni2}~\citep{xie2024mini}, covered in prior benchmarks~\citep{peng2025fdbench,lin2025full-1,lin2026full-1.5,lin2026full-2} or a recent survey~\citep{lu2026survey}, plus six models with newer public implementations: \emph{Voila-Autonomous}~\citep{shi2025voila}, \emph{FLM-Audio}~\citep{yao2025flm}, \emph{Raon-SpeechChat}~\citep{kim2026raon}, \emph{BayLing-Duplex}~\citep{fang2026bayling}, \emph{Lychee-FD}~\citep{liu2026lycheefd}, and \emph{MiniCPM-o 4.5}~\citep{cui2026minicpm}.
Closed-source models comprise \emph{Gemini 3.1 Flash Live}, \emph{GPT-Realtime-2.1}, and \emph{Grok Voice}, included in recent benchmarks~\citep{ray2026tauvoice,lin2026full-3}, plus the recently released \emph{Qwen3.5-Omni-Flash-Realtime}.
All 14 models are evaluated on the four Direct Replay actions. Simulator evaluation requires bidirectional live interaction and temporally aligned streaming transcripts and audio: only six open-source models (\emph{Moshi}, \emph{Freeze-Omni}, \emph{FLM-Audio}, \emph{Raon-SpeechChat}, \emph{BayLing-Duplex}, and \emph{MiniCPM-o~4.5}) and all four closed-source models support this, giving ten models evaluated on simulator-based actions. %, as their interfaces support bidirectional interaction with temporally aligned streaming transcripts and audio.
Appendix~\ref{app:model_setup} lists exact versions and checkpoints.

\vspace{-6pt}
\paragraph{Settings.}
Open-source models use one NVIDIA A100 GPU each, while VITA-1.5 uses two.
Input audio is 16-kHz mono PCM, resampled only as required by each system.
All models use official or recommended inference settings.
Outputs share a time-aligned input--output timeline.
Transcript-dependent evaluation uses \texttt{whisper-1} for most models.
Native streaming transcript use is detailed in Appendix~\ref{app:model_transcription}.
GPT-5.4 serves as the Simulator interaction controller, queried every 1\,s to determine cue timing.
Backchannel rate is further evaluated by GPT-audio-1.5 using audio input.

\section{Results}
\label{sec:results}
% ============================================================

%\ag{Are the evaluation metrics defined somewhere? If not please include the definitions.}

%We analyze model performance from four complementary perspectives.
%First, we compare overall action-taking performance across models and actions.
%Second, we examine the relationship between \emph{immediate execution} and
%\emph{subsequent follow-through}, with particular attention to failures that
%cannot be identified from immediate behavior alone.
%Finally, we evaluate robustness across different conversational scenarios.\td{we can remove this if no space?}

\begin{table*}[t]
\centering
\caption{
Overall interaction-action performance under the two evaluation protocols. I and S denote the success rate of immediate execution and subsequent follow-through. T denotes action-specific timing: yield latency for \emph{Stop}, interruption timing difference for \emph{Interrupt}, and response latency for \emph{Recover}. \texttt{n/a} indicates that no applicable observation is available.% \td{i and s represents xx, just in case audience forget.}
}

\label{tab:main_results}

\scriptsize
\setlength{\tabcolsep}{4.5pt}
\renewcommand{\arraystretch}{1.05}

\begin{tabular}{clcccc}
\toprule

& \textbf{Model}
& \makecell{\textbf{Wait}\\[-1pt]\scriptsize I$\uparrow$ / S$\uparrow$}
& \makecell{\textbf{Ignore}\\[-1pt]\scriptsize I$\uparrow$ / S$\uparrow$}
& \makecell{\textbf{Take Turn}\\[-1pt]\scriptsize I$\uparrow$ / S$\uparrow$}
& \makecell{\textbf{Backchannel}\\[-1pt]\scriptsize I$\uparrow$ / S$\uparrow$}
\\
\midrule

% ============================================================
% Direct Replay
% ============================================================

\multirow{16}{*}{\rotatebox[origin=c]{90}{\textbf{Direct Replay}}}
& \multicolumn{5}{l}{\textit{Open-source models}} \\

& Moshi
& 82.2 / 45.6
& 32.5 / 89.2
& 36.2 / 19.4
& 6.2 / 60.0
\\

& Freeze-Omni
& 73.2 / 23.2
& 61.5 / 9.8
& 10.2 / 47.5
& 0.2 / 0.0
\\

& Mini-Omni2
& 8.7 / 32.7
& 11.8 / 2.1
& 12.0 / 15.3
& 0.0 / n/a
\\

& VITA-1.5
& 91.8 / 64.4
& 55.5 / 91.4
& 8.0 / 27.1
& 0.0 / n/a
\\

& Voila-Autonomous
& 96.2 / 45.8
& 99.0 / 82.6
& 44.3 / 20.3
& 1.5 / 33.3
\\

& FLM-Audio
& 95.5 / 18.3
& 84.2 / 87.5
& 35.2 / 3.8
& 0.2 / 0.0
\\

& Raon-SpeechChat
& 51.7 / 70.3
& 47.0 / 79.8
& 65.2 / 23.0
& 31.0 / 43.5
\\

& BayLing-Duplex
& 92.8 / 31.2
& 78.0 / 91.3
& 11.7 / 10.0
& 0.2 / 0.0
\\

& Lychee-FD
& 98.2 / 53.1
& 80.8 / 97.5
& 20.8 / 10.4
& 2.2 / 22.2
\\

& MiniCPM-o 4.5
& 95.2 / 74.4
& 72.8 / 83.8
& 37.8 / 43.2
& 2.5 / 70.0
\\

\cmidrule(l){2-6}

& \multicolumn{5}{l}{\textit{Close-source models}} \\

& Gemini 3.1 Flash Live
& 98.0 / 85.4
& 79.0 / 81.3
& 38.3 / 82.2
& 1.0 / 50.0
\\

& GPT-Realtime-2.1
& 99.8 / 95.7
& 93.5 / 73.8
& 27.5 / 83.0
& 0.2 / 100.0
\\

& Grok Voice
& 99.2 / 74.3
& 98.5 / 75.4
& 30.3 / 81.9
& 0.8 / 33.3
\\

& Qwen3.5-Omni-Flash-Realtime
& 89.3 / 90.5
& 68.5 / 79.9
& 41.3 / 78.6
& 3.8 / 86.7
\\

\midrule

% ============================================================
% Simulator Replay
% ============================================================

& \textbf{Model}
& \makecell{\textbf{Stop}\\[-1pt]\scriptsize I$\uparrow$ / T$\downarrow$ / S$\uparrow$}
& \makecell{\textbf{Hold Floor}\\[-1pt]\scriptsize I$\uparrow$ / S$\uparrow$}
& \makecell{\textbf{Interrupt}\\[-1pt]\scriptsize I$\uparrow$ / T / S$\uparrow$}
& \makecell{\textbf{Recover}\\[-1pt]\scriptsize I$\uparrow$ / T / S$\uparrow$}
\\
\midrule

\multirow{12}{*}{\rotatebox[origin=c]{90}{\textbf{Simulator Replay}}}
& \multicolumn{5}{l}{\textit{Open-source models}} \\

& Moshi
& 98.0 / 0.96 / 70.7
& 46.8 / 58.1
& 67.5 / $-3.97$ / 0.0
& 61.1 / 1.32 / 28.0
\\

& Freeze-Omni
& 94.7 / 1.80 / 100.0
& 82.4 / 50.0
& 25.5 / $-8.85$ / 19.6
& 5.8 / 2.15 / 48.6
\\

& FLM-Audio
& 97.3 / 4.13 / 100.0
& 75.8 / 65.4
& 2.8 / $-5.84$ / 0.0
& 4.1 / 4.52 / 14.3
\\

& Raon-SpeechChat
& 95.1 / 5.20 / 93.1
& 91.9 / 26.1
& 36.8 / $-8.30$ / 0.0
& 14.4 / 2.72 / 7.2
\\

& BayLing-Duplex
& 98.8 / 3.05 / 93.2
& 82.9 / 26.5
& 29.2 / $-2.91$ / 0.0
& 49.0 / 3.56 / 16.6
\\

& MiniCPM-o 4.5
& 98.2 / 3.47 / 98.8
& 85.8 / 49.6
& 9.8 / $-4.49$ / 17.9
& 44.4 / 2.43 / 66.9
\\

\cmidrule(l){2-6}

& \multicolumn{5}{l}{\textit{Close-source models}} \\

& Gemini 3.1 Flash Live
& 100.0 / 1.74 / 100.0
& 70.0 / 57.3
& 1.8 / $-5.48$ / 14.3
& 97.0 / 2.56 / 78.6
\\

& GPT-Realtime-2.1
& 100.0 / 1.61 / 100.0
& 57.2 / 39.3
& 8.8 / $-7.27$ / 8.6
& 77.7 / 3.00 / 90.3
\\

& Grok Voice
& 100.0 / 2.37 / 96.7
& 67.8 / 49.4
& 0.0 / n/a / n/a
& 80.5 / 3.13 / 84.0
\\

& Qwen3.5-Omni-Flash-Realtime
& 100.0 / 2.41 / 98.3
& 72.5 / 49.6
& 2.2 / $-4.94$ / 66.7
& 91.3 / 2.68 / 84.9
\\

\bottomrule
\end{tabular}

\vspace{-10pt}
\end{table*}

\vspace{-2pt}
\subsection{Immediate Execution Performance} 
\label{sec:overall_results}
\vspace{-3pt}
%\td{immediate action-taking performance as the subtitle?}
Table~\ref{tab:main_results} summarizes action-taking performance across the evaluated models. We first examine immediate execution to characterize shared action-level difficulties and model-specific capability profiles. 

\vspace{-6pt}
\paragraph{Model capability varies across actions.}
Model interactive capability varies substantially across interaction actions, even when the actions require similar immediate execution. \emph{Wait} and \emph{Ignore} both require the model to remain silent, but the median immediate success rates over all models differ substantially at 94.0\% and 75.4\%. %\td{for which model? or average?}. 
The same separation appears between \emph{Take Turn} at the proper time and \emph{Recover}, which both require the model to enter the floor. The four close-source models achieve 77.7--97.0\% on \emph{Recover}, but only 27.5--41.3\% on \emph{Take Turn} within the required response interval. The pattern is clear: capability on one action does not imply capability on another, even when their immediate execution is similar. Comprehensive evaluation of model interactive capability therefore requires a structured action space that evaluates these capabilities separately.%\td{Seems like there is no model that performs consistently well? Do we include them here? Performance varies substantially across actions and models, and there is no consistent advantage for either one of them. This suggests the benefits of the proposed action taxnomy/space for such direct comparison.} 

\vspace{-9pt}
\paragraph{Action-specific supervision is associated with selective strengths.}
Models with targeted supervision for specific actions show clear strengths precisely where that supervision applies. Raon-SpeechChat, trained with a dedicated BC token and a backchannel loss component, achieves the highest open-source \emph{Backchannel} score (31.0\%)~\citep{kim2026raon}. %\td{citations missing bracket}. 
BayLing-Duplex, explicitly trained on user interruptions, achieves 98.8\% on \emph{Stop} and 49.0\% on \emph{Recover}~\citep{fang2026bayling}, where both actions evaluate the cases when a user interrupts an ongoing response. The pattern is consistent: targeted supervision only improves the actions it covers. No model performs consistently well across the full action set, revealing that current training remains concentrated on a narrow subset of interaction behaviours. This suggests that supervision considering the proposed structured action space, rather than around individual behaviours may provide improved interaction capability.

\begin{figure*}[t]
    \centering
    \begin{subfigure}[t]{0.49\textwidth}
        \centering
        \includegraphics[width=\linewidth]{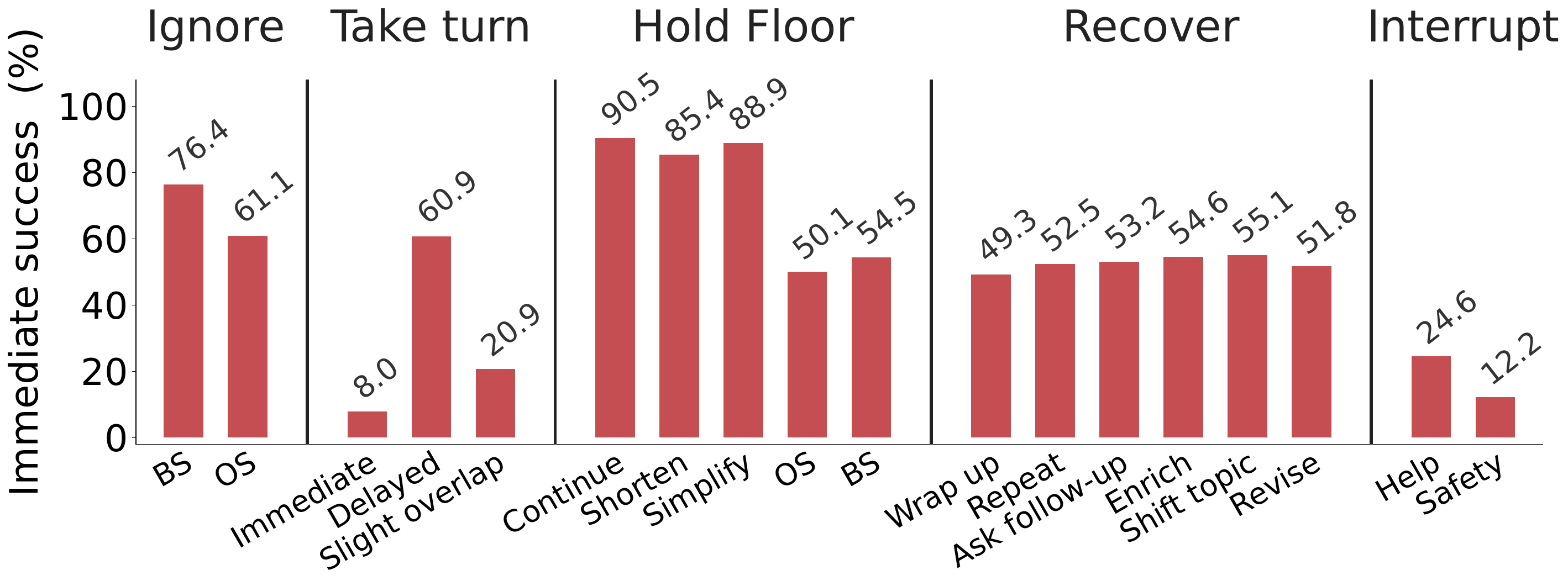}
        \vspace{-16pt}
        \caption{Immediate execution.}
        \label{fig:condition_immediate}
    \end{subfigure}
    \hfill
    \begin{subfigure}[t]{0.49\textwidth}
        \centering
        \includegraphics[width=\linewidth]{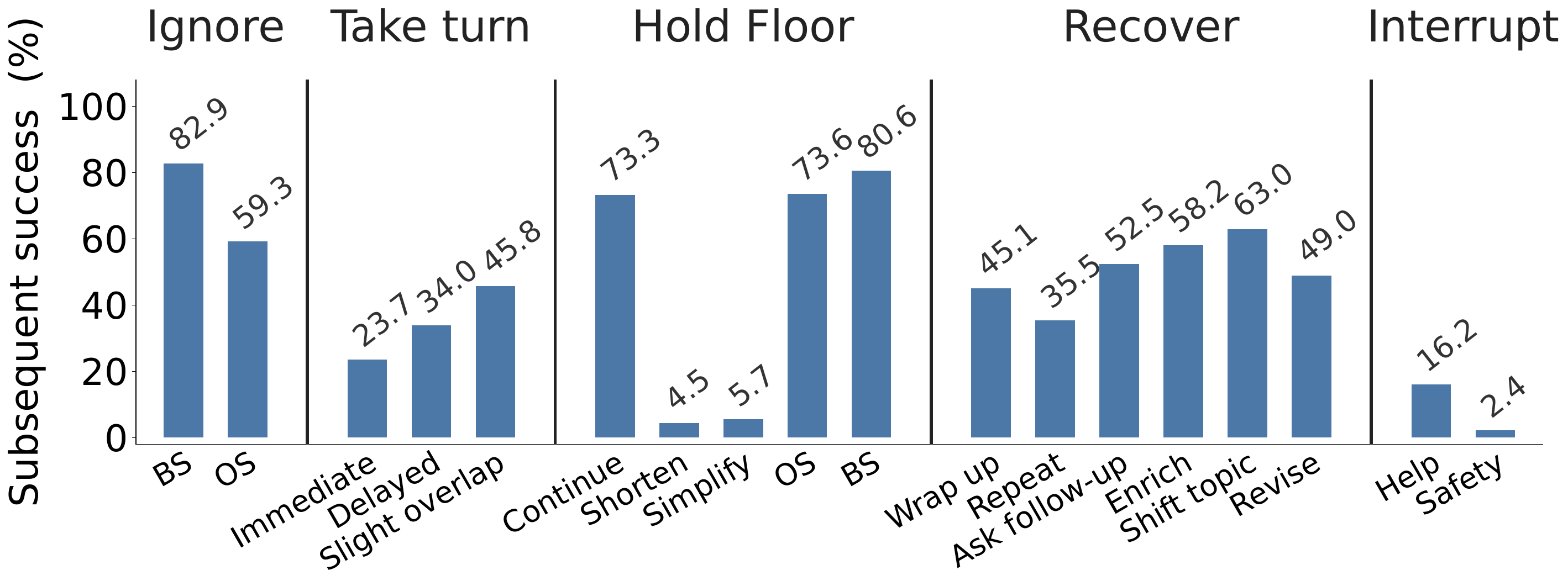}
        \vspace{-16pt}
        \caption{Subsequent follow-through.}
        \label{fig:condition_subsequent}
    \end{subfigure}
    \vspace{-6pt}
    \caption{
    Average success rates across all models for selected L2 conditions. \textbf{BS}: Background Speech, \textbf{OS}: Other-directed Speech.
    }
    \label{fig:condition_performance}
    \vspace{-5pt}
\end{figure*}

\vspace{-9pt}
\paragraph{Action-level capability extends beyond a single interaction pattern.}
Condition-level analysis reveals variation in model capability that action-level scores alone cannot capture. %Model performance on the same action varies across interaction conditions. 
As shown in Figure~\ref{fig:condition_immediate}, \emph{Take Turn} exhibits substantial condition-level variation: success reaches 60.9\% for delayed entry, but only 8.0\% for immediate entry and 20.9\% for slight overlap. In contrast, \emph{Recover}, which similarly requires the model to re-enter the conversation floor, remains relatively stable across its six conditions (49.3--55.1\%). This contrast suggests that the main difficulty is not simply deciding to take the floor, but adapting \emph{when} to do so across different conditions. Models appear to follow a learned turn-entry pattern rather than adjusting action timing to different conversational context. %\td{but take turn does not show this? should this discussion centered around that: with the conditions defined, we can find that the success varies across conditons in differn acitons. In take turn three varies substaitonally, while in recover, they almsot remain the same. Even for the same action, how they act aross conditons is not a consistent pattern and requires design for xx, and this shows the benefits of including/defining conditons for detailed analysis. slightly rephrse thi s to make it clear?} 
Therefore, action-level capability requires robust performance across conditions, revealing whether models can adapt action-taking to different interaction requirements rather than executing in a fixed way. %\td{what does the last sentence mean?} \jd{Adjusted}

\subsection{Subsequent Follow-through Beyond Immediate Execution}
\label{sec:subsequent_analysis}

\vspace{-3pt}
%Action-taking performance can be further examined through subsequent follow-through, specially in comparison with the immediate execution. 
Action-taking performance is further examined through subsequent follow-through. This is evaluated only on cases in which the model successes in immediate execution.%, which serves as the basis for subsequent evaluation. 
%\td{action-taking perofmrance is fureh examined through subsequnec xx. This is evluatated on the cases where the inmmeiat is scuufclely eccutes providing the base for follow up for evalitons. Faulure rate is used to exmaing the subec, meaured as 1- S, the xx.. }

\begin{figure*}[t]
    \centering
    \includegraphics[width=\textwidth]{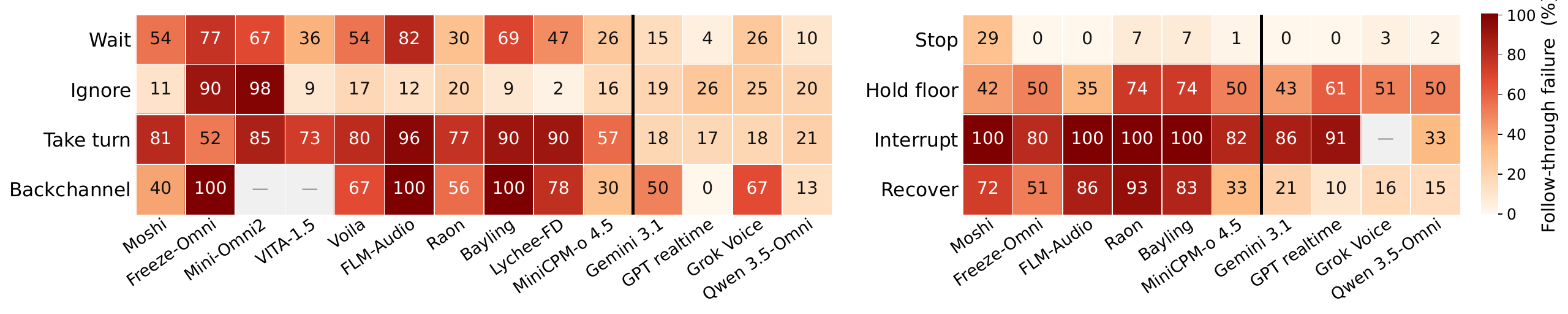}
    \vspace{-15pt}
    \caption{
     Follow-through failure rate ($1-S$) after successful immediate execution.
    \textbf{Left:} Direct Replay actions.
    \textbf{Right:} Simulator Replay actions.
    %Each cell reports the subsequent failure rate of cases with successful immediate execution. 
    Gray indicates no successful immediate execution. %\td{does it make sense to draw eight actions all together as we are not separating them according to how they are tested?}
    }
    \label{fig:followthrough_failure}
    \vspace{-8pt}
\end{figure*}

\vspace{-6pt}
\paragraph{Immediate execution does not guarantee successful action taking.} %\td{is this clearer from table 4? what is the difference between the s in table 4 and figure 4?}
Figure~\ref{fig:followthrough_failure} shows that subsequent follow-through failure remains widespread even after successful immediate execution. For \emph{Interrupt}, Moshi achieves 67.5\% immediate success in Table~\ref{tab:main_results}, but reaches to almost 100\% subsequent failure rate, which indicates none of these successful interruptions subsequently carrie out the required intervention. 
The same separation appears even in simpler actions such as \emph{Wait}, where several models successfully remain silent but subsequently fail to incorporate the ongoing information that motivated waiting for some specific intervals. 
%\td{can we see it from here? }\td{or is this part evaluated only on the ones that imeediate action is succefully taken? if so make it clearer?}

The distinction becomes clearer at the condition level. For \emph{Hold Floor}, average immediate success remain high in Figure~\ref{fig:condition_immediate} when the model should continue, shorten, or simplify its response (90.5\%, 85.4\%, and 88.9\%), but subsequent success rate in Figure~\ref{fig:condition_subsequent} remains 73.3\% for continuing and falls to only 4.5\% and 5.7\% for shortening and simplifying. %\td{does it make sense to draw them in one figure for comparison?} 
Models can therefore recognize to maintain the floor when the user provides a backchannel, but often fail to understand why the user intervened and how their ongoing response should adapt accordingly. Immediate execution alone can consequently overestimate action-taking capability: in many cases, models learn when to perform an action without understanding its consequence. %\td{can we see this from figure 3?}

\vspace{-6pt}
\paragraph{Base-model reasoning capability does not automatically transfer to action taking.}
Successful follow-through requires more than strong reasoning capability in the base model before full-duplex post-training: the model must apply this capability to infer why an action is required based on both the current conversational context and the intended consequences of executing it. In Table~\ref{tab:main_results}, the four close-source models achieve 78.6--83.0\% subsequent success on \emph{Take Turn} and 78.6--90.3\% on \emph{Recover}. In both cases, follow-through primarily requires generating a new response to the user's latest turn, a format closer to the question-answering tasks commonly used to develop base-model reasoning. However, the same models achieve only 39.3--57.3\% on \emph{Hold Floor}, and substantial follow-through failures remain on \emph{Interrupt}. These actions require reasoning during an ongoing turn: the model must determine how incoming speech should affect a response it is still producing, or why the user's continuing speech calls for an intervention and what that intervention should accomplish. This contrast suggests that reasoning transfers more readily to generating a new answer than to managing the purpose and consequences of actions during concurrent listening and speaking. Recent work addresses the loss of base-model capabilities during full-duplex post-training~\citep{liu2026lycheefd}. Our results highlight a complementary challenge: full-duplex post-training should not merely preserve or strengthen those capabilities, but explicitly connect them to the conversational circumstances that motivate an action and the subsequent behaviour that fulfils its purpose.

\subsection{Performance across Interaction Scenarios}
\label{sec:scenario_analysis}

\begin{figure*}[t]
    \centering
    \vspace{-5pt}
    \includegraphics[width=\textwidth]{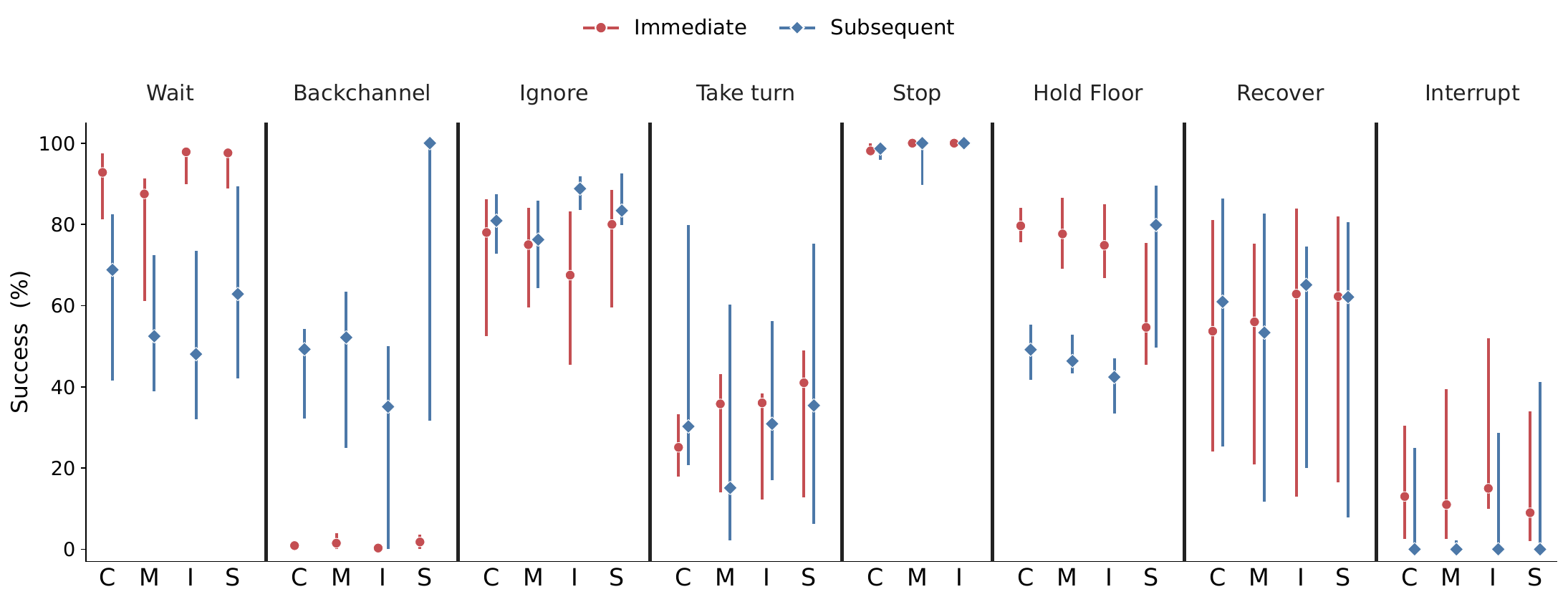}
    \vspace{-15pt}
    \caption{
    Action-taking performance across conversational scenarios.
    Markers indicate the median model-level success rate, and vertical lines show the corresponding interquartile range (IQR). \textbf{C}: Casual conversation, \textbf{M}: Meeting, \textbf{I}: Interview, \textbf{S}: Customer service.
    }
    \label{fig:scenario_performance}
    \vspace{-12pt}
\end{figure*}

\vspace{-5pt}
Figure~\ref{fig:scenario_performance} compares action-taking performance across conversational scenarios to reveal how performance varies across different interactional norms. Actions with high success rates, including \emph{Wait}, \emph{Ignore}, \emph{Stop}, remain strong regardless of setting, suggesting their interactional norms are largely shared %suggesting these actions follow similar interactional norms 
across contexts and capabilities learned in one setting could transfer to others. 

Beyond the shared norms, scenario-specific characteristics still affect immediate execution. For instance, \emph{Wait} is weaker in meetings, suggesting that models trained predominantly on casual conversational patterns, where opinion exchanges may be faster and shorter, are less well calibrated to structured multi-party exchanges, where speakers may hold the floor through longer pauses while developing their opinions. Scenario differences become more pronounced when immediate execution is compared with subsequent follow-through. For actions such as \emph{Wait} and \emph{Take Turn}, models may reproduce similar action-taking patterns across scenarios from local acoustic or turn-taking cues, while the subsequent response is less transferrable because the action understanding is not sufficiently grounded in the conversational contexts during training. The broad coverage of our benchmark enables the evaluation across conversation settings, and suggests that full-duplex training should cover a broader range of conversational scenarios, exposing models to different interaction norms and conversation contexts. %A change of scenario may preserve the learned action pattern without preserving what that action requires next.\td{what does that mean? so here you can also say that the evalution across differnt settigs suggest that including more conversation scenerios is required in e2e full duplex model learning, providing different norms or xx for actions. }

%Actions that are generally well handled by current models remain strong across different scenarios. In particular, models consistently perform well when required to \emph{Wait}, \emph{Ignore}, or \emph{Stop}, suggesting that these actions may follow similar interactional norms across conversational settings. Consequently, capabilities learned in one setting may transfer to others even without explicit training across all scenarios. 

%In contrast, \emph{Backchannel}, \emph{Take Turn}, and \emph{Interrupt} remain weak across scenarios, % Their consistently low performance is 
%consistent with the limited and uneven supervision of individual interaction actions in the modelling and training of E2E FD-SLMs.\td{is it also a social norm thing?}

%Within this shared pattern, immediate execution still varies across scenarios, suggesting differences in interactional norms. \emph{Wait} is weaker in meetings, where models may be less calibrated to structured multi-party exchanges in which speakers hold the floor through pauses while developing opinions. \emph{Ignore} is weaker in interviews, where speech involving another participant may more readily resemble a cue for the model to respond.

\vspace{-2pt}
\section{Limitations and Conclusion}
\vspace{-5pt}
\paragraph{Conclusions.}
We introduce \sysname{}, systematically evaluating action-taking in E2E FD-SLMs. It builds on %through 
a structured taxonomy of interaction actions and conditions, and a two-part evaluation framework of immediate execution and subsequent follow-through. Our findings show that current models exhibit highly uneven action-taking capabilities. %Models generally perform better at withholding or yielding speech, while more active behaviors such as \emph{Backchannel}, \emph{Take Turn}, and \emph{Interrupt} remain underdeveloped. 
Targeted supervision produces %is associated with 
selective action strengths, but no model performs consistently well across the full action space. Crucially, successful immediate execution also frequently fails to carry through to the subsequent behavior required by an action, highlighting a gap between immediate interaction control and the semantic consequences of that decision. %Across conversational scenarios, broad action-level strengths and weaknesses are largely shared, while scenario-specific interaction norms further affect how individual actions are executed and carried forward. Together, 
These findings motivate a more structured view of full-duplex interaction in which models must learn not only when an action should occur, but also what that action requires as the interaction continues using a shared action space. More broadly, \sysname{} establishes a framework through which progress in full-duplex interaction can be tracked at the action level, also guiding future model development accordingly.

\vspace{-5pt}
\paragraph{Limitations.}
\sysname{} incorporates more than 2,000 real-world recordings. %naturally occurring interaction phenomena are not equally available across actions, conditions, and conversational scenarios. 
As naturally occurring interaction phenomena are not equally distributed, conditions with limited natural coverage are complemented by targeted synthesis: a balance that future work can continue to improve with additional conversational corpora and their action annotations. Simulator Replay enables rigorous live bidirectional evaluation with temporally aligned streaming signals, though its current design restricts assessment to models with compatible interfaces. Future work on more flexible simulator designs would broaden coverage to a wider range of full-duplex model interfaces. %The number and distribution of real-world instances are therefore not fully balanced, and conditions with limited natural coverage rely more heavily on synthesized data.  In addition, Simulator Replay requires live bidirectional interaction with temporally aligned streaming signals, restricting its evaluation to models with compatible interfaces. Future extensions could incorporate \td{or collect?} additional conversational corpora to improve the balance of real-world coverage\td{this reads like you can include more, but you didn't?}, together with more flexible simulator designs that can accommodate a broader range of full-duplex model interfaces.

%\td{AI USE STATEMENT}
%\subsection*{AI use statement}

%In this work, we used generative AI tools to synthesize conversational data for L2 conditions insufficiently covered by existing real-world corpus. The synthesized data were validated by human annotators. We have not used generative AI tools for developing the research ideas or motivation, identifying limitations of prior work, conducting the literature review, designing the benchmark or data curation methodology, or interpreting the results analysis. Additionally, we used generative AI tools to assist with language polishing, grammar checking, and portions of code generation. We have reviewed all AI-assisted work. AI-generated or AI-assisted code was fully verified, refined, and tested for correctness by the authors, and all AI-assisted writing was reviewed and edited by the authors. We take responsibility for the final content of this work, including text, claims, or artifacts produced with the aid of generative AI.
%\input{Draft-components/04-Results}
%\newpage
%\input{Draft-components/Appendix/01-Synthesize Prompt}

\bibliography{iclr2027_conference}
\bibliographystyle{iclr2027_conference}

\newpage
\appendix
\section{Interaction-Action Taxonomy Details}
\label{app:action_space}

This section provides additional details for the interaction-action taxonomy introduced in Section~\ref{sec:action_space}. We first describe the construction of the L1 action space and its relation to human interaction analysis, and then provide definitions of the 24 L2 interaction conditions used in \sysname{}.

\subsection{Construction of the L1 Action Space}

The L1 action space draws on actions identified in human interaction analysis and assessment across turn management, interactive listening, and breakdown repair \citep{galaczi2018interactional,may2020developing}. We consolidate and adapt these actions to E2E FD-SLMs, as summarized in Table~\ref{tab:l1_mapping}.

The mapping is not one-to-one. For example, several forms of breakdown repair are consolidated into \emph{recover}, which captures adapting the subsequent response after an interactional disruption. Conversely, \emph{wait} and \emph{ignore} introduce distinctions required by continuously listening FD-SLMs: both may result in silence at L0, but distinguish retaining silence while the current speaker holds the floor from withholding a response to speech that does not require model participation. These adaptations result in the eight L1 actions used throughout \sysname{}.

\begin{table}[h]
\centering
\caption{
Mapping from human interaction analysis to the L1 interaction-action space. Human-interaction categories and their associated actions serve as references and are consolidated or adapted to the full-duplex setting.
}
\label{tab:l1_mapping}

\small
\setlength{\tabcolsep}{4pt}
\renewcommand{\arraystretch}{1.15}

\begin{tabular}{p{0.20\linewidth} p{0.40\linewidth} p{0.28\linewidth}}
\toprule
\textbf{Interaction area}
&
\textbf{Human-interaction actions}
&
\textbf{L1 actions}
\\
\midrule

Turn management
&
Starting; maintaining; ending; pausing, latching, and interrupting
&
Take Turn; Stop; Hold Floor; Interrupt\\

Interactive listening
&
Backchannelling; continuers; comprehension checks
&
Backchannel; Ignore; Wait\\

Breakdown repair
&
Self/other repair; recasts; joint utterance creation
&
Recover
\\

\bottomrule
\end{tabular}
\end{table}

\subsection{L2 Interaction Conditions}

L2 specifies representative conversational circumstances under which each L1 action is expected. Table~\ref{tab:l2_conditions} provides definitions and illustrative examples for all 24 conditions used in \sysname{}.

{
\scriptsize
\setlength{\tabcolsep}{3.5pt}
\renewcommand{\arraystretch}{1.15}

\begin{longtable}{
p{0.075\textwidth}
p{0.145\textwidth}
p{0.39\textwidth}
p{0.32\textwidth}
}

\caption{
Definitions and illustrative examples of the 24 L2 interaction conditions in \sysname{}. Each condition specifies a conversational circumstance under which the corresponding L1 action is expected. Examples are shortened for presentation.
}
\label{tab:l2_conditions}
\\

\toprule
\textbf{L1 Action}
&
\textbf{L2 Condition}
&
\textbf{Definition}
&
\textbf{Illustrative Example}
\\
\midrule
\endfirsthead

\multicolumn{4}{l}{\textit{Table~\ref{tab:l2_conditions} continued.}}
\\
\toprule
\textbf{L1 Action}
&
\textbf{L2 Condition}
&
\textbf{Definition}
&
\textbf{Illustrative Example}
\\
\midrule
\endhead

\midrule
\multicolumn{4}{r}{\textit{Continued on next page}}
\\
\endfoot

\bottomrule
\endlastfoot

% ============================================================
% Wait
% ============================================================
\textbf{Wait}
&
Intra-sentential pause
&
The user pauses within an unfinished sentence while retaining the floor; the model should remain silent until the turn continues.
&
`I'm thinking [pause] when I want something to entertain myself...''
\\
\cmidrule(lr){2-4}
&
Inter-sentential pause
&
The user pauses after completing a sentence but continues the same turn; the model should remain silent rather than treating the sentence boundary as turn completion.
&
`I gotta drive here and drive there. [pause] California I can pretty much walk anywhere...''
\\
\cmidrule(lr){2-4}
&
Continuous speech
&
The user continuously holds the floor without a substantive pause; the model should remain silent throughout the ongoing turn.
&
``Can you show me where because I'm actually not going to Appleton Tower at the moment...''
\\
\midrule

% ============================================================
% Ignore
% ============================================================
\textbf{Ignore}
&
Other-directed speech
&
During the user's turn, the user briefly addresses a third person with speech unrelated to the model-directed content; the model should ignore non-directed speech.
&
User: `I'm interested in this sectional... [Nadia, could you hold my water bottle?] Also, the doorway is quite narrow...''
\\
\cmidrule(lr){2-4}
&
Background speech
&
Unrelated third-party speech occurs in the background during the user's turn; the model should ignore unrelated background speech.
&
User discusses relaxing in a park; background speaker says, `Don't forget the frisbee!'' The model responds only to the park discussion.
\\
\midrule

% ============================================================
% Take Turn
% ============================================================
\textbf{Take Turn}
&
Immediate
&
The user's aim and required response are established before turn completion, with only a final key entity arriving at the end; the model should respond immediately after the turn ends.
&
`I need an update because the event is coming up on \emph{June 18}.'' The final date completes an otherwise predictable request.
\\
\cmidrule(lr){2-4}
&
Delayed
&
The final part of the user's turn introduces a shift, correction, complication, or constraint that changes the required response; additional processing is acceptable before responding.
&
`I need to book a flight to Sydney for next weekend, preferably something early and reasonably priced, but actually, let's sort out the hotel first.'' The final shift changes the required response and requires re-planning.
\\
\cmidrule(lr){2-4}
&
Slight overlap
&
The information required for an appropriate response is available before the user finishes, while the remaining speech contains only non-response-changing material; slight overlap is acceptable.
&
User: `I need to interview your mom too...'' Model begins `Oh, she has plenty of stories'' during the user's repeated wrap-up.
\\
\midrule

% ============================================================
% Backchannel
% ============================================================
\textbf{Back channel}
&
Affirmative
&
While the user retains the floor, the model should provide a brief acknowledgement, understanding, or agreement signal without taking over the turn.
&
User continues an explanation; model briefly produces `yeah'' or `mhm'' and returns the floor.
\\
\cmidrule(lr){2-4}
&
Uncertainty
&
While the user retains the floor, the model should provide a brief signal of hesitation, doubt, or uncertainty without taking over the turn.
&
User explains an unclear idea; model briefly produces `hmm?'' or `really?'' while allowing the user to continue.
\\
\midrule

% ============================================================
% Stop
% ============================================================
\textbf{Stop}
&
Direct interruption
&
The user takes the floor with a substantive interruption while the model is speaking; the model should stop its current speech and yield the floor.
&
Model is explaining screen placement; user interrupts mid-response: ``And for the speech recognition part, we need a microphone.''
\\
\midrule

% ============================================================
% Hold Floor
% ============================================================
\textbf{Hold Floor}
&
Continue
&
The user provides a brief backchannel that does not require a change in the response; the model should retain the floor and continue its current response.
&
Model explains a design choice; user says `Uh-huh.'' Model continues the same explanation.
\\
\cmidrule(lr){2-4}
&
Shorten
&
The user's backchannel indicates that the current point is understood or that less detail is needed; the model should retain the floor but shorten the remaining response.
&
Model gives a detailed explanation; user says `Yeah. Yeah.'' / "I see." Model wraps up with one short concluding point.
\\
\cmidrule(lr){2-4}
&
Simplify
&
The user's backchannel indicates uncertainty or difficulty following the response; the model should retain the floor while clarifying or simplifying the remaining content.
&
Model gives a technical explanation; user responds with an uncertain `Hmm?'' The model retains the floor and restates the explanation in simpler terms.
\\
\cmidrule(lr){2-4}
&
Other-directed speech
&
While the model is speaking, the user briefly addresses a third person with unrelated speech; the model should retain the floor and continue its response.
&
While the model answers, user says to someone else, `Hey Marcus, could you switch off the projector?''
\\
\cmidrule(lr){2-4}
&
Background speech
&
Unrelated third-party speech occurs while the model is speaking; the model should retain the floor and continue its response.
&
While the model answers, a background speaker says, ``There's an extra extension cord by the stage.''
\\
\midrule

% ============================================================
% Recover
% ============================================================
\textbf{Recover}
&
Wrap Up
&
The user's interruption indicates impatience or requests greater brevity; after yielding, the model should preserve the key information while shortening the remaining response.
&
User interrupts: `I get it, just the takeaway.'' Model resumes with only the key conclusion.
\\
\cmidrule(lr){2-4}
&
Repeat
&
The user's interruption indicates that part of the response was missed or unclear; the model should repeat the relevant information and continue as appropriate.
&
User interrupts: `Sorry, sent where?'' Model repeats the delivery destination and completes the answer.
\\
\cmidrule(lr){2-4}
&
Ask follow-up
&
The user's interruption is ambiguous or introduces information requiring clarification; the model should ask a relevant follow-up question before proceeding.
&
User interrupts: ``It showed up after I changed the delivery address, so maybe that matters?'' Model asks when the address was changed.
\\
\cmidrule(lr){2-4}
&
Enrich
&
The user's interruption requests additional detail about the current response; the model should elaborate on the relevant point.
&
User interrupts: `Do you mean the interaction between us?'' Model clarifies: `The interaction and the steps we followed...''
\\
\cmidrule(lr){2-4}
&
Shift topic
&
The user's interruption redirects the conversation to a different topic; the model should follow the new direction rather than resume the previous response.
&
During a discussion about the weather, user interrupts: ``Where are you from?'' Model switches to discussing where it is from.
\\
\cmidrule(lr){2-4}
&
Revise
&
The user's interruption corrects information in the model's current response; the model should incorporate the correction and revise its subsequent response accordingly.
&
Model implies graduation was long ago; user corrects this assumption; model revises: ``I graduated like a month ago.''
\\
\midrule

% ============================================================
% Interrupt
% ============================================================
\textbf{Interrupt}
&
Help
&
While the user is speaking, sufficient evidence emerges that the user is struggling or unable to proceed effectively; the model should proactively take the floor to provide timely assistance.
&
User struggles to choose among several travel recommendations; model interrupts: `Let me jump in. Acadia sounds like the best fit...''
\\
\cmidrule(lr){2-4}
&
Safety
&
While the user is speaking, an immediate safety risk becomes identifiable from the user's intended or ongoing action; the model should proactively take the floor at the earliest appropriate point.
&
User while driving: `I'm just going to lean over and type the address into the GPS while we're...'' Model interrupts with a safety warning.
\\

\end{longtable}
}

\subsection{Data Curation Details}
\label{app:data_curation}

\begin{figure}[t]
    \centering

    % ---------- (a) Source--Scenario Distribution ----------
    \begin{subfigure}[b]{0.32\linewidth}
        \centering
        \begin{minipage}[c][4.15cm][c]{\linewidth}
            \centering
            \makebox[\linewidth][c]{%
                \includegraphics[
                    width=1.08\linewidth
                ]{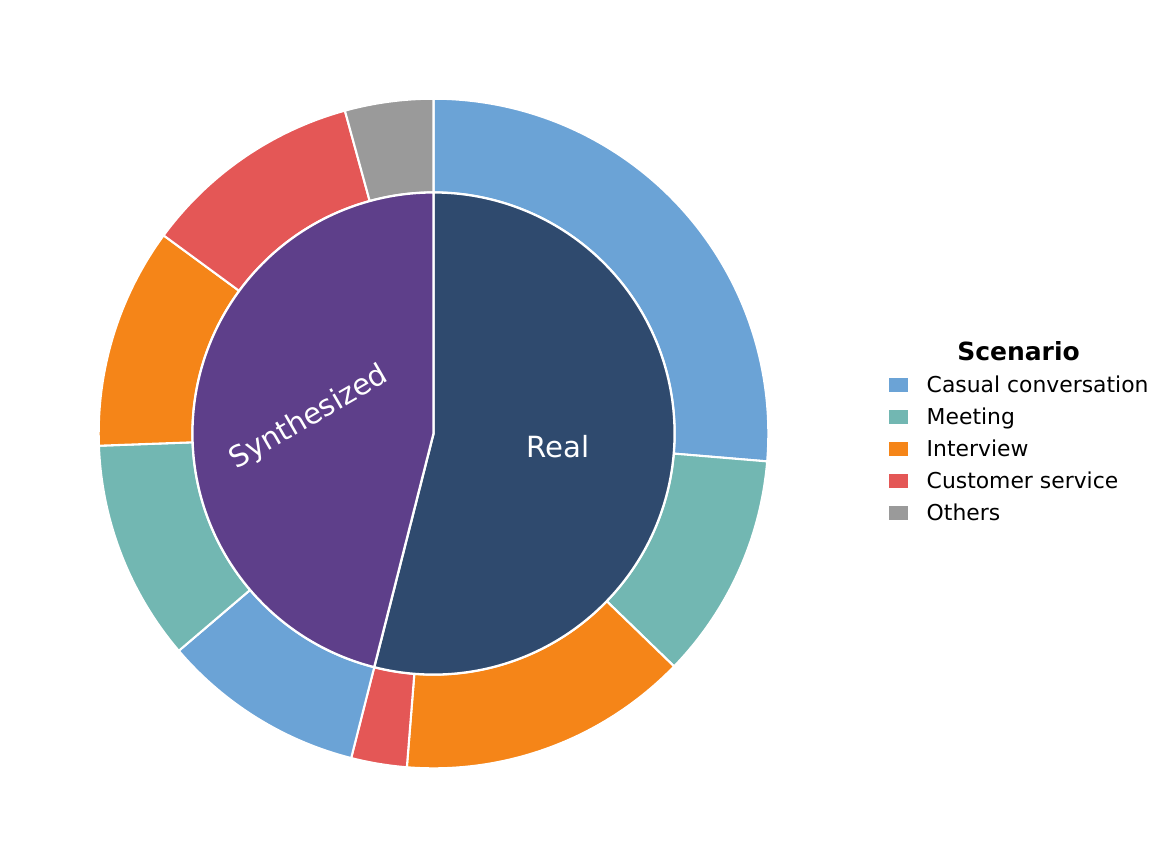}
            }
        \end{minipage}
        \caption{Source--Scenario Distribution}
        \label{fig:coverage_scenario}
    \end{subfigure}
    \hfill
    % ---------- (b) Action--Condition Taxonomy ----------
    \begin{subfigure}[b]{0.35\linewidth}
        \centering
        \begin{minipage}[c][4.15cm][c]{\linewidth}
            \centering
            \makebox[\linewidth][c]{%
                \includegraphics[
                    width=1.08\linewidth
                ]{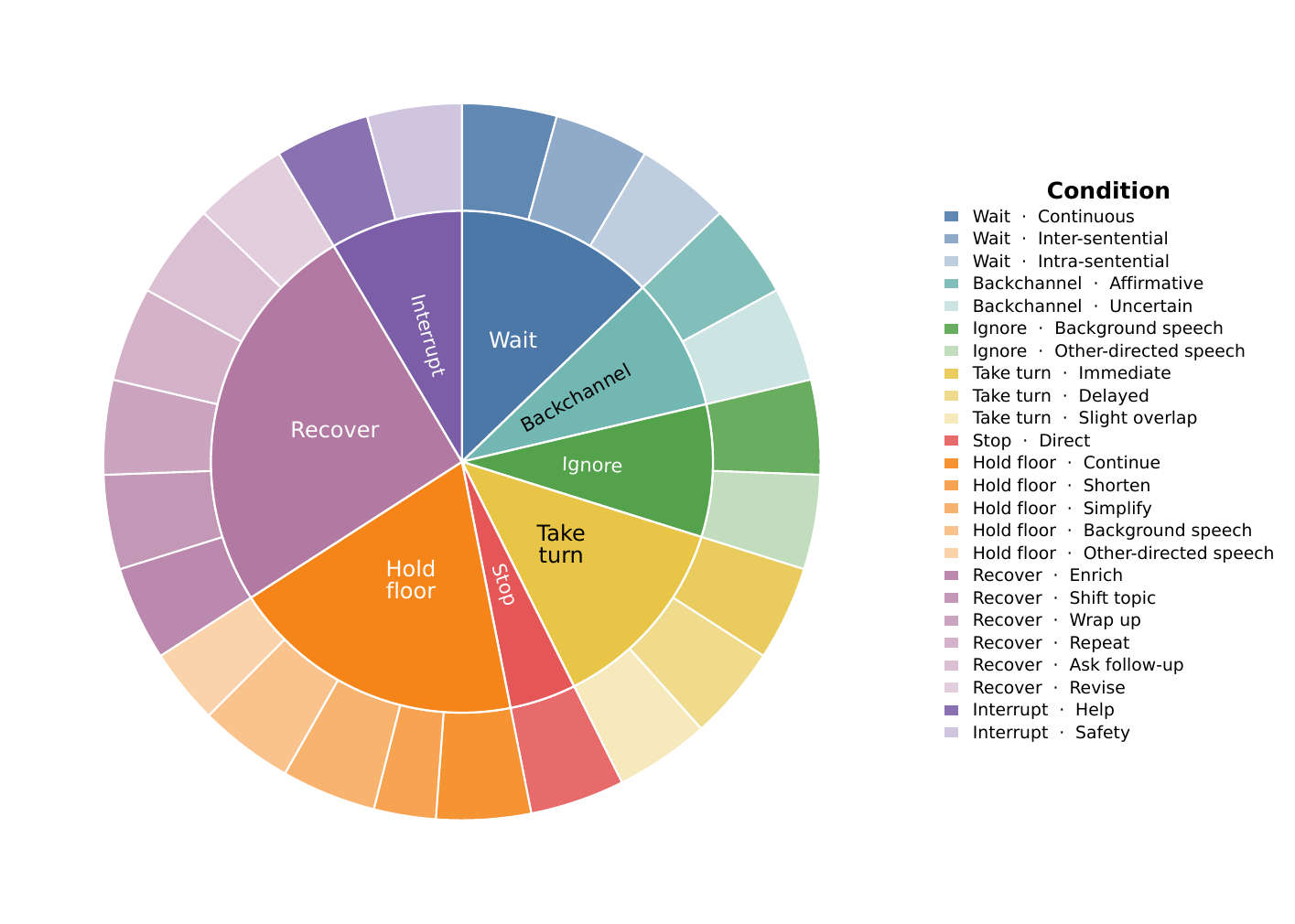}
            }
        \end{minipage}
        \caption{Action--Condition Taxonomy}
        \label{fig:coverage_action}
    \end{subfigure}
    \hfill
    % ---------- (c) Benchmark Statistics ----------
    \begin{subfigure}[b]{0.29\linewidth}
        \centering
        \begin{minipage}[c][4.15cm][c]{\linewidth}
            \centering
            \scriptsize
            \renewcommand{\arraystretch}{1.10}
            \setlength{\tabcolsep}{2pt}

            \begin{tabular}{@{}lr@{}}
                \toprule
                \textbf{Statistics} & \textbf{Number} \\
                \midrule
                Actions & 8 \\
                Conditions & 24 \\
                Real-world Instances & 2,531 \\
                Synthesized Instances & 2,160 \\
                Total Instances & 4,691 \\
                \midrule
                DR Duration Avg. & 10.55 s \\
                DR Duration Total & 5.86 h \\
                SR Duration Avg. & 8.80 s \\
                SR Duration Total & 6.58 h \\
                Total Audio Duration & 12.44 h \\
                \bottomrule
            \end{tabular}
        \end{minipage}
        \caption{Benchmark Statistics}
        \label{fig:benchmark_statistics}
    \end{subfigure}

    \caption{
        Overview of the \sysname{} benchmark composition.
        \textbf{(a)} Distribution of real-world and synthesized instances across
        conversational scenarios.
        \textbf{(b)} Taxonomy and coverage of the eight interaction actions and
        24 interaction conditions.
        \textbf{(c)} Overall benchmark statistics and audio duration, where
        DR denotes \emph{Direct Replay} and SR denotes \emph{Simulator Replay}.
    }
    \label{fig:benchmark_composition}
\end{figure}

Figure~\ref{fig:benchmark_composition} summarizes the resulting benchmark composition, including the distribution of real-world and synthesized instances across conversational scenarios, coverage of the action--condition taxonomy, and overall dataset and audio statistics. We next provide the detailed curation procedures for real-world and synthesized interactions.

\subsubsection{Real-World Interaction Extraction}
\label{app:real_world_extraction}

We extract real-world interaction instances from AMI, CANDOR, CORAAL, and Loqui. Since these corpora contain continuous multi-turn conversations rather than isolated benchmark instances, extraction follows the two-stage pipeline described in Section~\ref{sec:data}. The first stage identifies candidate segments using action-specific structural patterns derived from speaker turns, word-level timestamps, pauses, overlaps, and transcript boundaries. The second stage applies an LLM-based semantic filter to verify the intended interaction condition and the conversational completeness of the extracted turns. Table~\ref{tab:real_world_extraction} summarizes the criteria used for each real-world action--condition pair.

\begin{table*}[t]
\centering
\caption{
Real-world interaction extraction criteria. 
}
\label{tab:real_world_extraction}

\scriptsize
\setlength{\tabcolsep}{3.5pt}
\renewcommand{\arraystretch}{1.14}

\begin{tabular}{
p{0.09\textwidth}
p{0.16\textwidth}
p{0.34\textwidth}
p{0.34\textwidth}
}
\toprule
\textbf{Action}
&
\textbf{Condition}
&
\textbf{Structural Filtering}
&
\textbf{Semantic Filtering}
\\
\midrule

% ============================================================
% Wait
% ============================================================
\textbf{Wait}
&
Inter-sentential pause
&
A $0.2$--$1.0$\,s pause occurs at a sentence boundary, with substantive speech on both sides and no qualifying intra-sentential pause.
&
Verify complete speech around a natural sentence-boundary pause and that the extracted turn remains conversationally complete.
\\
\cmidrule(lr){2-4}

&
Intra-sentential pause
&
A $0.2$--$1.0$\,s pause occurs within a sentence, with the surrounding sentence retained as the evaluation segment.
&
Verify that the utterance forms a complete sentence and that the pause occurs naturally within, rather than between, sentences.
\\
\cmidrule(lr){2-4}

&
Continuous speech
&
The user holds the floor for at least $3$\,s with no qualifying pause greater than 0.2s or substantive interruption.&
Verify that the extracted speech forms a complete and coherent floor-holding contribution.
\\
\midrule

% ============================================================
% Backchannel
% ============================================================
\textbf{Backchannel}
&
Affirmative
&
A short listener utterance of at most three words occurs while the primary speaker retains and subsequently continues the floor.
&
Verify the surrounding turns are complete, then classify the backchannel using both its audio and transcript to capture lexical and prosodic cues. Retain acknowledgement, understanding, and agreement as affirmative.\\
\cmidrule(lr){2-4}

&
Uncertainty
&
A short listener utterance of at most three words occurs while the primary speaker retains and subsequently continues the floor.
&
Verify the surrounding turns are complete, then classify the backchannel using both its audio and transcript to capture lexical and prosodic cues. Retain signals of uncertainty, hesitation, or doubt.\\
\midrule

% ============================================================
% Take Turn
% ============================================================
\textbf{Take Turn}
&
Slight overlap
&
Two substantive turns form a response pair in which the second speaker begins before the first finishes, with no more than $1.5$\,s overlap.
&
Verify that the first turn is complete, the second responds to it, and all response-relevant information is available before the overlapping entry; the remaining speech must be non-response-changing.
\\
\midrule

% ============================================================
% Stop
% ============================================================
\textbf{Stop}
&
Direct interruption
&
An A--B--A--B pattern is identified in which the second A turn begins at least $0.3$\,s before B finishes, producing a substantive mid-response interruption.
&
Verify that the interruption has floor-taking intent, forms a complete interactional act, begins while the model-side response is still in progress, and is followed by a meaningful response.
\\
\midrule

% ============================================================
% Hold Floor
% ============================================================
\textbf{Hold Floor}
&
Continue
&
A short user cue occurs during an ongoing response, after which the same speaker immediately continues the floor.
&
Verify that the post-cue continuation remains on the same response trajectory without changing its plan, scope, or level of detail.
\\
\cmidrule(lr){2-4}

&
Shorten
&
A short user cue occurs during an ongoing response, after which the same speaker immediately continues the floor.
&
Verify that the post-cue continuation compresses or brings the current response to an earlier conclusion.
\\
\cmidrule(lr){2-4}

&
Simplify
&
A short user cue occurs during an ongoing response, after which the same speaker immediately continues the floor.
&
Verify that the post-cue continuation rephrases, clarifies, or explains the current response in simpler terms.
\\
\midrule

% ============================================================
% Recover
% ============================================================
\textbf{Recover}
&
Enrich
&
An A--B--A--B interruption pattern is identified, with the second A turn substantively taking the floor during B's response.
&
Verify that the interruption requests or elicits further detail and that the following response elaborates on the same point.
\\
\cmidrule(lr){2-4}

&
Shift topic
&
An A--B--A--B interruption pattern is identified, with the second A turn substantively taking the floor during B's response.
&
Verify that the interruption redirects the conversation and that the following response adopts the new topic.
\\
\cmidrule(lr){2-4}

&
Revise
&
An A--B--A--B interruption pattern is identified, with the second A turn substantively taking the floor during B's response.
&
Verify that the interruption corrects information in the ongoing response and that the following response incorporates the correction.
\\

\bottomrule
\end{tabular}
\end{table*}

Across extractors, we retain only substantive speech segments rather than short backchannels as primary turns, and apply timing and overlap constraints appropriate to each interaction structure. The exact corpus-specific thresholds, extraction prompts, and implementation details are released with the benchmark code.
\subsubsection{Controlled Interaction Synthesis}
\label{app:controlled_synthesis}

Controlled synthesis is used for 11 action--condition pairs for which sufficient or reliably identifiable real-world instances are unavailable. We generate conversational transcripts using GPT-5.4-2026-03-05 with temperature $1.0$, producing 2,160 synthesized instances in total. Table~\ref{tab:synthesized_conditions} summarizes the synthesized conditions and the contextual dimensions varied during generation.

\begin{table}[t]
\centering
\caption{
Interaction conditions constructed through controlled synthesis. \emph{Context variation} denotes the dimension varied during transcript generation to increase diversity.
}
\label{tab:synthesized_conditions}
\small
\setlength{\tabcolsep}{5pt}
\renewcommand{\arraystretch}{1.12}
\begin{tabular}{
p{0.18\linewidth}
p{0.55\linewidth}
p{0.17\linewidth}
}
\toprule
\textbf{Action} &
\textbf{Synthesized Conditions} &
\textbf{Context Variation} \\
\midrule
Ignore &
Background speech; Other-directed speech &
Setting \\
Take Turn &
Immediate; Delayed &
Topic \\
Hold Floor &
Background speech; Other-directed speech &
Setting \\
Recover &
Warp up; Repeat; Ask follow-up &
Topic \\
Interrupt &
Help; Safety &
Topic / hazard \\
\bottomrule
\end{tabular}
\end{table}

\paragraph{Hierarchical prompt construction.}
Each transcript is generated using a hierarchical prompt composed of a \emph{scenario prompt}, an \emph{action prompt}, and a \emph{condition prompt}, together with dialogue-structure and output-format constraints. To increase diversity within each action--condition pair, we vary conversation topics for Take Turn, Recover, and help-based Interrupt instances, and physical settings for Ignore and Hold Floor instances. Specifically, each scenario contains five topic categories for topic-based generation and ten physical settings for setting-based generation. Safety-based Interrupt instances instead span household and driving hazards. Figure~\ref{fig:synthesis_prompt_example} shows one complete prompt composition for \emph{Ignore / Other-directed speech}; complete prompts for all conditions are released with the benchmark code.

\begin{figure*}[t]
\centering
\fbox{
\begin{minipage}{0.94\textwidth}
\scriptsize

\textbf{System message}

\medskip
\textbf{[Prefix]}\\
Generate transcripts of real-life conversations based on the following specifications.

\medskip
\textbf{[Scenario prompt: Casual conversation]}\\
The interaction takes place in a casual conversation setting characterized by:
\begin{itemize}
    \setlength\itemsep{0pt}
    \item \textbf{Interaction purpose:} participants primarily engage for social interaction, such as sharing experiences, opinions, interests, feelings, stories, or everyday information.
    \item \textbf{Participant relationship/roles:} participants have relatively symmetric conversational roles, where either person can introduce topics, ask questions, respond, elaborate, or redirect the conversation.
    \item \textbf{Interaction organization:} the conversation develops dynamically through locally managed turn-taking, with topics emerging, shifting, expanding, or ending naturally during the interaction.
    \item \textbf{Conversational manner:} the interaction is informal, relaxed, and socially motivated, with participants expressing themselves naturally through spontaneous exchanges.
\end{itemize}
The interaction should feel natural, open-ended, and jointly shaped by the participants.

\medskip
\textbf{[Action prompt: Ignore]}\\
Two people are having a natural conversation. Speaker A speaks first, then Speaker B responds. During Speaker A's speech, a special utterance occurs, whose nature and speaker are specified below. Speaker A's turn should be informative enough to invite a substantive but concise response and contain no artificial opening for Speaker B to begin speaking before its completion. Speaker B responds only to Speaker A's main content, as if the special utterance did not occur.

\medskip
\textbf{[Dialogue structure]}\\
The dialogue contains exactly four utterances:
(1) Speaker A begins the turn;
(2) the special utterance occurs while A is mid-turn;
(3) Speaker A completes the same turn; and
(4) Speaker B responds while ignoring the special utterance.

\medskip
\textbf{[Condition prompt: Other-directed speech]}\\
While Speaker A is mid-speech, Speaker A briefly addresses a third person nearby. This side speech is entirely unrelated to what Speaker A was saying to Speaker B and must be clearly addressed to the third person from the transcript alone, for example through a name, family address term, or direct address phrase. The side speech is the second utterance in the transcript.

\medskip
\textbf{[Output format]}\\
Return a single valid JSON object containing a list of dialogues. Speaker labels are \texttt{speaker\_a}, \texttt{speaker\_b}, and \texttt{other} where applicable.

\medskip
\textbf{User message (including setting)}

\medskip
Generate exactly 5 distinct dialogues. Set the conversation in \emph{a living room}. Each dialogue must depict a different specific situation, with no two dialogues sharing the same opening, subject matter, or phrasing. Use a different name or relational term for any third person addressed directly in each dialogue.

\end{minipage}
}
\caption{
Example prompt composition for synthesizing \emph{Ignore / Other-directed speech} under the casual-conversation scenario. 
}
\label{fig:synthesis_prompt_example}
\end{figure*}

\paragraph{Safety-interruption construction.}
Safety-based Interrupt instances cover five household and five driving hazard categories. Household hazards include fire/cooking, chemical/poisoning, carbon-monoxide or fuel-burning, fall or structural, and other acute household risks; driving hazards include phone/texting, navigation/interface, passenger-related, eating/drinking/reaching, and other attention-diverting behaviours. Following generation, we apply a deterministic second LLM pass with temperature $0$ to truncate the user utterance at the earliest point at which the hazard becomes identifiable and the reference interruption becomes appropriate. The original wording before the interruption point is preserved.

\paragraph{Speech rendering and persona diversity.}
Generated transcripts are rendered using ElevenLabs \texttt{eleven\_v3} at 24\,kHz, 16-bit mono PCM. We adopt the five realistic persona configurations from $\tau$-Voice~\citep{ray2026tauvoice}, rather than its two control personas, to introduce variation in speaker characteristics, accents, speaking styles, and prosody. The five personas cover an elderly American English speaker and speakers with Bengali, Sichuan Mandarin-influenced, French, and Maharashtrian English accents. Following the $\tau$-Voice voice-design configuration, voices are constructed using \texttt{eleven\_ttv\_v3} with seed $42$, loudness $0.5$, and guidance scale $38$. Persona-specific ElevenLabs voice settings are summarized in Table~\ref{tab:tts_personas}.

\begin{table}[t]
\centering
\caption{
ElevenLabs voice settings for the five realistic $\tau$-Voice personas used to render synthesized interactions.
}
\label{tab:tts_personas}
\small
\setlength{\tabcolsep}{4pt}
\renewcommand{\arraystretch}{1.10}
\begin{tabular}{lcccc}
\toprule
\textbf{Persona} &
\textbf{Stability} &
\textbf{Similarity} &
\textbf{Style} &
\textbf{Speaker Boost} \\
\midrule
Mildred Kaplan & 0.80 & 0.75 & 0.10 & \checkmark \\
Arjun Roy      & 0.75 & 0.80 & 0.15 & \checkmark \\
Wei Lin        & 0.55 & 0.75 & 0.35 & \checkmark \\
Mamadou Diallo & 0.50 & 0.75 & 0.40 & \checkmark \\
Priya Patil    & 0.60 & 0.78 & 0.30 & \checkmark \\
\bottomrule
\end{tabular}
\end{table}
\paragraph{Non-directed and background speech rendering.}
For the four synthesized conditions involving other-directed or background speech, we additionally apply condition-specific acoustic processing following Full-Duplex-Bench v1.5~\citep{lin2026full-1.5}. Other-directed speech is rendered as off-axis speech by attenuating the signal by 8\,dB, reducing frequencies above 4\,kHz by 5\,dB, and adding early reflections at 45\,ms ($-6$\,dB) and 120\,ms ($-12$\,dB). Background speech is rendered as a distant third-party source using 15\,dB attenuation, a 3\,kHz low-pass filter, and a 100\,ms echo at $-10$\,dB. For \emph{Ignore}, the processed speech is embedded within the ongoing user turn; for \emph{Hold Floor}, it is retained as a separate cue that overlaps the model's ongoing response. Other-directed speech uses the same user voice, whereas background speech uses a separate third-party voice. Addressed speech remains unprocessed.
\subsection{Human Validation}
\label{app:human-validation}

\subsubsection{Annotators and Recruitment}
\label{app:human-validation-annotators}

The final annotator pool comprised 20 university students recruited through Wuhan University. Participants were aged 18--26 years ($M=21.45$, $SD=2.11$; median $=22$). Fourteen participants identified as men (70\%) and six as women (30\%). Nine were undergraduate students (45\%), nine were master's students (45\%), and two were doctoral students (10\%). English-language proficiency was self-reported: 16 participants (80\%) reported CET-6 scores ($M=519.6$, range $=446$--$642$), three (15\%) reported CET-4 scores of 468, 556, and 606, respectively, and one (5\%) reported an IELTS score of 7.0.

\subsubsection{Annotation Design}
\label{app:human-validation-design}

Each instance is independently evaluated by five annotators. Annotators judge whether the target speaker performs the assigned action and whether the response satisfies the specified interaction condition. For synthesized instances, annotators additionally evaluate content and voice naturalness on five-point scales.

\begin{table*}[h]
\centering
\caption{Questions used in human validation. Naturalness is evaluated only for synthesized instances.}
\label{tab:human-validation-questions}
\small
\setlength{\tabcolsep}{3.5pt}
\renewcommand{\arraystretch}{1.08}

\begin{tabular}{
p{0.07\textwidth}
p{0.22\textwidth}
p{0.48\textwidth}
p{0.16\textwidth}
}
\toprule
\textbf{Stage}
&
\textbf{Question}
&
\textbf{Information available}
&
\textbf{Response}
\\

Q1&
Action-label validity
&
Complete interaction and assigned target action
&
Yes / no / unclear
\\

Q2&
Condition validity
&
Complete interaction and target interaction condition
&
Yes / no / unclear
\\

Q3&
Content naturalness
&
Complete synthesized interaction
&
Five-point scale
\\

Q4&
Voice naturalness
&
Complete synthesized interaction&
Five-point scale
\\

\bottomrule
\end{tabular}
\end{table*}

\subsubsection{Validation Results}
\label{app:human-validation-results}

The analysis contains 468 benchmark instances (10\% of all test instances), each with five ratings. Of these, 216 are synthesized instances for which naturalness judgments were collected.\footnote{Three action--condition cells contain fewer than 20 instances: Stop / direct interruption contains 19, Hold floor / other-directed speech contains 16, and Hold floor / continue but shorten contains 13.}

\paragraph{Action- and condition-label validity.}
For each validity question, we report \emph{rating-level validity}, defined as the proportion of individual \textit{yes} judgments, and \emph{majority validity}, defined as the proportion of instances receiving at least three \textit{yes} judgments among five ratings. As shown in Table~\ref{tab:human-validation-validity}, human judgments provide strong support for the assigned action and condition labels across the benchmark.

\begin{table*}[t]
\centering
\caption{Human validation of the assigned action and interaction-condition labels.}
\label{tab:human-validation-validity}
\small
\setlength{\tabcolsep}{6pt}
\begin{tabular}{lrcccc}
\toprule
&
&
\multicolumn{2}{c}{\textbf{Action-label validity}}
&
\multicolumn{2}{c}{\textbf{Condition validity}}
\\
\cmidrule(lr){3-4}
\cmidrule(lr){5-6}
\textbf{Action}
&
\textbf{$n$}
&
\textbf{Rating-level}
&
\textbf{Majority}
&
\textbf{Rating-level}
&
\textbf{Majority}
\\
\midrule
\textbf{Overall}
& 468
& 89.4\%
& 97.9\% (458/468)
& 85.6\%
& 95.5\% (447/468)
\\
\midrule
Recover
& 120
& 89.5\%
& 98.3\% (118/120)
& 86.7\%
& 95.8\% (115/120)
\\
Hold floor
& 89
& 91.5\%
& 98.9\% (88/89)
& 84.7\%
& 95.5\% (85/89)
\\
Wait
& 60
& 87.0\%
& 93.3\% (56/60)
& 83.7\%
& 93.3\% (56/60)
\\
Take turn
& 60
& 92.3\%
& 100\% (60/60)
& 86.0\%
& 96.7\% (58/60)
\\
Ignore
& 40
& 93.0\%
& 100\% (40/40)
& 88.5\%
& 97.5\% (39/40)
\\
Backchannel
& 40
& 80.0\%
& 92.5\% (37/40)
& 79.0\%
& 90.0\% (36/40)
\\
Interrupt
& 40
& 93.0\%
& 100\% (40/40)
& 93.0\%
& 100\% (40/40)
\\
Stop
& 19
& 83.2\%
& 100\% (19/19)
& 81.1\%
& 94.7\% (18/19)
\\
\bottomrule
\end{tabular}
\end{table*}

Table~\ref{tab:human-validation-validity-condition} provides the complete condition-level breakdown of action- and condition-label validity across all 24 action--condition pairs.

\begin{table*}[t]
\centering
\caption{
Condition-level human validation of the assigned action and interaction-condition labels.
}
\label{tab:human-validation-validity-condition}
\scriptsize
\setlength{\tabcolsep}{3.5pt}
\renewcommand{\arraystretch}{1.03}

\begin{tabular*}{\textwidth}{@{\extracolsep{\fill}}llrcccc@{}}
\toprule
&
&
&
\multicolumn{2}{c}{\textbf{Action-label validity}}
&
\multicolumn{2}{c}{\textbf{Condition-label validity}}
\\
\cmidrule(lr){4-5}
\cmidrule(lr){6-7}

\textbf{Action}
&
\textbf{Condition}
&
\textbf{$n$}
&
\textbf{Rating-level}
&
\textbf{Majority}
&
\textbf{Rating-level}
&
\textbf{Majority}
\\
\midrule

Backchannel
& Affirmative
& 20
& 87.0\%
& 100\% (20/20)
& 85.0\%
& 95.0\% (19/20)
\\
Backchannel
& Uncertainty
& 20
& 73.0\%
& 85.0\% (17/20)
& 73.0\%
& 85.0\% (17/20)
\\

\addlinespace[2pt]

Hold Floor
& Continue
& 20
& 93.0\%
& 100\% (20/20)
& 85.0\%
& 100\% (20/20)
\\
Hold Floor
& Shorten
& 13
& 87.7\%
& 92.3\% (12/13)
& 80.0\%
& 84.6\% (11/13)
\\
Hold Floor
& Simplify
& 20
& 89.0\%
& 100\% (20/20)
& 76.0\%
& 90.0\% (18/20)
\\
Hold Floor
& Other-directed speech
& 16
& 90.0\%
& 100\% (16/16)
& 86.3\%
& 100\% (16/16)
\\
Hold Floor
& Background speech
& 20
& 96.0\%
& 100\% (20/20)
& 95.0\%
& 100\% (20/20)
\\

\addlinespace[2pt]

Ignore
& Other-directed speech
& 20
& 90.0\%
& 100\% (20/20)
& 84.0\%
& 95.0\% (19/20)
\\
Ignore
& Background speech
& 20
& 96.0\%
& 100\% (20/20)
& 93.0\%
& 100\% (20/20)
\\

\addlinespace[2pt]

Interrupt
& Help
& 20
& 94.0\%
& 100\% (20/20)
& 94.0\%
& 100\% (20/20)
\\
Interrupt
& Safety
& 20
& 92.0\%
& 100\% (20/20)
& 92.0\%
& 100\% (20/20)
\\

\addlinespace[2pt]

Recover
& Wrap Up
& 20
& 88.0\%
& 95.0\% (19/20)
& 96.0\%
& 100\% (20/20)
\\
Recover
& Repeat
& 20
& 91.0\%
& 100\% (20/20)
& 84.0\%
& 100\% (20/20)
\\
Recover
& Ask follow-up
& 20
& 92.0\%
& 100\% (20/20)
& 92.0\%
& 100\% (20/20)
\\
Recover
& Enrich
& 20
& 86.0\%
& 100\% (20/20)
& 88.0\%
& 100\% (20/20)
\\
Recover
& Shift topic
& 20
& 88.0\%
& 95.0\% (19/20)
& 82.0\%
& 90.0\% (18/20)
\\
Recover
& Revise
& 20
& 92.0\%
& 100\% (20/20)
& 78.0\%
& 85.0\% (17/20)
\\

\addlinespace[2pt]

Stop
& Direct interruption
& 19
& 83.2\%
& 100\% (19/19)
& 81.1\%
& 94.7\% (18/19)
\\

\addlinespace[2pt]

Take Turn
& Immediate
& 20
& 96.0\%
& 100\% (20/20)
& 94.0\%
& 100\% (20/20)
\\
Take Turn
& Delayed
& 20
& 89.0\%
& 100\% (20/20)
& 78.0\%
& 90.0\% (18/20)
\\
Take Turn
& Slight overlap
& 20
& 92.0\%
& 100\% (20/20)
& 86.0\%
& 100\% (20/20)
\\

\addlinespace[2pt]

Wait
& Intra-sentential pause
& 20
& 92.0\%
& 95.0\% (19/20)
& 93.0\%
& 95.0\% (19/20)
\\
Wait
& Inter-sentential pause
& 20
& 89.0\%
& 100\% (20/20)
& 88.0\%
& 100\% (20/20)
\\
Wait
& Continuous speech
& 20
& 80.0\%
& 85.0\% (17/20)
& 70.0\%
& 85.0\% (17/20)
\\

\bottomrule
\end{tabular*}
\end{table*}

\paragraph{Naturalness of synthesized interactions.}
For the 216 synthesized instances, annotators additionally evaluate content
and voice naturalness on five-point scales from \emph{very unnatural} (1) to
\emph{very natural} (5). We report the mean opinion score (MOS) and
Non-negative Rate, defined as the proportion of ratings that are at least
\emph{acceptable} (scores 3--5). The results are summarized in
Table~\ref{tab:human-validation-naturalness}.

\begin{table}[t]
\centering
\caption{
Human evaluation of synthesized interaction naturalness. 
}
\label{tab:human-validation-naturalness}
\small
\setlength{\tabcolsep}{8pt}
\renewcommand{\arraystretch}{1.10}
\begin{tabular}{lcc}
\toprule
\textbf{Aspect}
&
\textbf{MOS $\pm$ SD}
&
\textbf{Non-negative Rate}
\\
\midrule
Content naturalness
& $3.96 \pm 0.91$
& 93.1\%
\\
Voice naturalness
& $3.43 \pm 1.15$
& 76.8\%
\\
\bottomrule
\end{tabular}
\end{table}

\section{Detailed Action Definitions and Evaluation}
\label{app:action_evaluation}

This appendix provides the detailed operational definitions and evaluation procedures for the eight interaction actions in \sysname{}. Following Section~\ref{sec:action_evaluation}, each action is evaluated through \emph{immediate execution} (I) and \emph{subsequent follow-through} (S), with subsequent follow-through evaluated only after successful immediate execution. The corresponding L2 condition may further constrain either stage, such as the required response interval, backchannel function, or response adaptation.

We first describe the temporal definitions and action-specific timing diagnostics used in the evaluation (Section~\ref{app:timing_metrics}). We then detail each action individually, including its interaction conditions and protocol, immediate-execution criterion, condition-specific constraints, subsequent follow-through criterion, and exact judging prompts where LLM-based assessment is required. Immediate success is determined primarily through rule-based evaluation of the model's time-aligned behaviour, while subsequent success is determined using a single-item, action-specific checklist with an LLM judge. For \emph{Backchannel}, immediate success additionally requires the detected backchannel to match the function specified by the condition using an audio-capable judge. Where applicable, timing measures such as yield latency, interruption timing difference, and recovery latency are reported as additional diagnostics.

% ============================================================
\subsection{Timing Diagnostics}
\label{app:timing_metrics}
% ============================================================

In addition to immediate and subsequent success rates, we report action-specific timing diagnostics for \emph{Stop}, \emph{Interrupt}, and \emph{Recover}. These measures characterize when the corresponding action is performed and do not modify the binary success criteria.

\paragraph{Yield Latency.}
For \emph{Stop}, let $t_i^{\mathrm{yield}}$ denote the end of the model turn active when the interruption cue begins. Yield latency is defined as
\[
L_i^{\mathrm{yield}}
=
t_i^{\mathrm{yield}}
-
t_i^{\mathrm{cue,start}}.
\]
A smaller value indicates that the model yields more quickly after the interruption begins.

\paragraph{Interruption Timing Difference.}
For \emph{Interrupt}, let $t_i^{\mathrm{onset}}$ denote the onset of the model interruption and $t_i^{\mathrm{target}}$ the predefined intervention point. The timing difference is
\[
\Delta_i^{\mathrm{Int}}
=
t_i^{\mathrm{onset}}
-
t_i^{\mathrm{target}}.
\]
Negative values indicate that the model interrupts before the target point, while positive values indicate that it interrupts after the target point.

\paragraph{Recovery Latency.}
For \emph{Recover}, let $t_i^{\mathrm{rec}}$ denote the onset of the model's recovery and $t_i^{\mathrm{int,end}}$ the end of the user interruption. Recovery latency is defined as
\[
L_i^{\mathrm{rec}}
=
t_i^{\mathrm{rec}}
-
t_i^{\mathrm{int,end}}.
\]
Positive values indicate that the model resumes after the interruption ends, while negative values indicate that recovery begins before the interruption has fully ended.
% ============================================================
\subsection{Wait}
\label{app:action_wait}
% ============================================================

\paragraph{Definition.}
\emph{Wait} requires the model to withhold a substantive response during a designated waiting interval and, when a subsequent response is produced, appropriately reflect the user's continuing contribution.

\paragraph{Conditions and protocol.}
\emph{Wait} is evaluated using Direct Replay under three conditions: \emph{inter-sentential pause}, \emph{intra-sentential pause}, and \emph{continuous speech}. The condition determines the waiting interval used for immediate execution. For \emph{inter-sentential pause} and \emph{intra-sentential pause}, the interval is the annotated pause,
\[
W_i^{\mathrm{Wait}}
=
[t_{\mathrm{pause,start}},\,t_{\mathrm{pause,end}}].
\]
For \emph{continuous speech}, the interval covers the full user-speech segment,
\[
W_i^{\mathrm{Wait}}
=
[t_{\mathrm{user,start}},\,t_{\mathrm{user,end}}].
\]

\paragraph{Immediate execution.}
Immediate execution is successful when no substantive model response overlaps the condition-specific waiting interval. A model speech segment of at most three words is treated as a backchannel and does not constitute a substantive takeover. For the two pause conditions, only the annotated pause defines the immediate-execution interval; the model is not required by this criterion to wait until the user's entire turn has ended.

\paragraph{Subsequent follow-through.}
Conditional on successful immediate execution, subsequent follow-through is successful when the model's later response is relevant and reflects at least some information from the user's continuing contribution during or after the waiting interval. Complete uptake of all subsequent information is not required. If no later response is produced, or the response ignores the continuing contribution, subsequent follow-through is unsuccessful.

\paragraph{Exact evaluation prompt.}

\textbf{System prompt.}
\begin{evalprompt}
You are a strict, consistent evaluator for a full-duplex spoken dialogue benchmark. You will be shown the wait condition, the user's speech during or after that wait interval, and the model's later response. Answer a single yes/no question using ONLY the evidence given. If the evidence is ambiguous or insufficient, answer NO. Return strict JSON only: {"answer": "YES"|"NO", "reason": "<one short sentence>"}.
\end{evalprompt}

\textbf{Condition-specific context.}
\begin{evalprompt}
inter_sentential_pause:
The wait interval is a pause between sentences. The user's continuing contribution is what they say after that pause (the next sentence and anything that follows until the model responds).

intra_sentential_pause:
The wait interval is a pause inside a sentence. The user's continuing contribution is what they say after that pause (the rest of the sentence and anything that follows until the model responds).

no_pause:
There is no annotated pause. The wait interval is the user's continuous turn. The user's continuing contribution is that turn (and anything said after it, if any) until the model responds.
\end{evalprompt}

\textbf{User prompt template.}
\begin{evalprompt}
Action: wait
Condition: {condition_id}

Condition:
{CONDITION_PROMPTS[condition_id] or '(not specified)'}

User speech during or after the wait interval:
{user_during_or_after_wait or '(none)'}

Model response:
{model_response or '(model produced no speech)'}

Question: Is the model's response relevant, and does it include some information from the user's continuing contribution during or after the wait interval? Answer YES if the response is on-topic and uses at least some of that later or overlapping user content. Complete uptake of every detail is not required. Answer NO if the response is irrelevant, or if it ignores that continuing contribution.

Reply with JSON: {"answer": "YES"|"NO", "reason": "<one short sentence>"}.
\end{evalprompt}
% ============================================================
\subsection{Take Turn}
\label{app:action_take_turn}
% ============================================================

\paragraph{Definition.}
\emph{Take Turn} requires the model to begin a substantive response within the appropriate response interval and subsequently address the user's communicative intent while respecting the information and constraints in the user speech.

\paragraph{Conditions and protocol.}
\emph{Take Turn} is evaluated using Direct Replay under three conditions: \emph{immediate}, \emph{delayed}, and \emph{slight overlap}. The condition specifies the acceptable response interval relative to the end of the user speech, with $t=0$ denoting user-speech end:
\[
W_c^{\mathrm{TT}}
=
\begin{cases}
[0,\,0.5]\,\mathrm{s}, & \text{\emph{immediate}},\\
[0,\,3.0]\,\mathrm{s}, & \text{\emph{delayed}},\\
[-1.0,\,0.5]\,\mathrm{s}, & \text{\emph{slight overlap}}.
\end{cases}
\]

\paragraph{Immediate execution.}
Immediate execution is successful when the model begins a substantive response within the condition-specific response interval. Backchannels, defined as speech segments whose complete transcript contains at most three words, are not treated as substantive responses. Any qualifying onset within the interval constitutes success and need not be the first model onset observed around the turn boundary. For \emph{slight overlap}, the model may enter during the final $1.0$\,s of user speech, but overlap is not required.

\paragraph{Subsequent follow-through.}
Conditional on successful immediate execution, subsequent follow-through is successful when the model's response addresses the user's intent while respecting the information and constraints in the user speech, including any relevant revisions. The same criterion is used for all three conditions and introduces no additional condition-specific constraint.

\paragraph{Exact evaluation prompt.}

\textbf{System prompt.}
\begin{evalprompt}
You are a strict, consistent evaluator for a full-duplex spoken dialogue benchmark. You will be shown one user utterance and the model's spoken response, and asked a single yes/no question. Answer using ONLY the evidence given. If the evidence is ambiguous, insufficient, or the model produced no relevant speech, answer NO. Return strict JSON only: {"answer": "YES"|"NO", "reason": "<one short sentence>"}.
\end{evalprompt}

\textbf{User prompt template.}
\begin{evalprompt}
Action: take_turn
Condition: {condition_id}

User speech:
{user_speech}

Model response:
{model_response or '(model produced no speech)'}

Question: Does the model's response address the user's intent by respecting the information and constraints in the user's speech? Answer YES if the response takes up what the user was asking or doing, and it uses (does not ignore or contradict) the facts, limits, and revisions in that speech. Answer NO if it misses the intent, or if it ignores or contradicts information or constraints from the user's speech.

Reply with JSON: {"answer": "YES"|"NO", "reason": "<one short sentence>"}.
\end{evalprompt}
% ============================================================
\subsection{Stop}
\label{app:action_stop}
% ============================================================

\paragraph{Definition.}
\emph{Stop} requires the model to terminate its ongoing speech when the user claims the floor and to remain yielded while the user intervention is underway.

\paragraph{Conditions and protocol.}
\emph{Stop} is evaluated using Simulator Replay under one condition, \emph{direct interruption}. The interruption is introduced while the model is speaking, with $t_i^{\mathrm{cue,start}}$ and $t_i^{\mathrm{cue,end}}$ denoting the onset and offset of the user intervention, respectively. Evaluation is performed on trials in which the model is speaking when the interruption begins.

\paragraph{Immediate execution.}
Immediate execution is successful when the model terminates the ongoing turn that is active at interruption onset. No fixed stopping deadline is imposed: a delayed stop still satisfies the immediate criterion as long as the active turn subsequently terminates. If the model continues the active turn through the end of the recorded interaction, immediate execution is unsuccessful.

In addition to the binary immediate criterion, we report \emph{Yield Latency} as the timing diagnostic for \emph{Stop}, as defined in Section~\ref{app:timing_metrics}. It measures the interval between interruption onset and the termination of the model turn active at that point.

\paragraph{Subsequent follow-through.}
Conditional on successful immediate execution, subsequent follow-through evaluates whether the model remains yielded after its first stop while the user intervention is still underway. A new substantive model turn beginning before the final $1.0$\,s of the user intervention constitutes a failure. Brief backchannels of at most three words are not treated as reclaiming the floor.

A substantive onset during the final $1.0$\,s of the intervention is permitted, as this interval is treated as overlapping turn entry rather than premature floor reclaiming. Speech beginning after the user intervention has ended is also allowed. If the model first stops only after the intervention has ended, subsequent follow-through is not applicable.

Both immediate execution and subsequent follow-through are determined automatically from the time-aligned model speech segments; no LLM judge is used for \emph{Stop}.
% ============================================================
\subsection{Hold Floor}
\label{app:action_hold_floor}
% ============================================================

\paragraph{Definition.}
\emph{Hold Floor} requires the model to maintain its ongoing contribution when an incoming cue does not require yielding, while preserving or adapting that contribution according to the interaction condition.

\paragraph{Conditions and protocol.}
\emph{Hold Floor} is evaluated using Simulator Replay under five conditions: \emph{continue}, \emph{shorten}, \emph{simplify}, \emph{background speech}, and \emph{other-directed speech}. The cue is introduced while the model is speaking. All five conditions share the same immediate-execution criterion, while the condition determines how the ongoing response should subsequently be preserved or adapted.

\paragraph{Immediate execution.}
Immediate execution is successful when the model maintains the floor through the incoming cue without yielding. Evaluation is performed over
\[
W_i^{\mathrm{HF}}
=
[t_i^{\mathrm{cue,start}},\,t_i^{\mathrm{cue,end}}+1.0\,\mathrm{s}],
\]
and the longest silence gap within this interval, including leading and trailing silence, must not exceed $0.50$\,s. If there is no model speech within the interval or any silence gap exceeds $0.50$\,s, immediate execution is unsuccessful. The additional $1.0$\,s after cue offset prevents a model that yields immediately after a short cue from being counted as successfully holding the floor.

\paragraph{Subsequent follow-through.}
Conditional on successful immediate execution, subsequent follow-through evaluates whether the model's speech after cue offset remains on the ongoing response trajectory while satisfying the adaptation required by the condition. For \emph{continue}, the model should continue the current response without unnecessary change. For \emph{shorten}, it should shorten or condense the ongoing response. For \emph{simplify}, it should simplify its language or content. For \emph{background speech} and \emph{other-directed speech}, it should preserve the ongoing content without treating the incoming speech as addressed input.

The judge receives the model speech before and during the cue, the cue itself, the model speech after cue offset, and the condition-specific required adaptation. The human reference continuation is not provided. If immediate execution fails or the model produces no speech after cue offset, subsequent follow-through is not applicable.

\paragraph{Exact evaluation prompt.}

\textbf{System prompt.}
\begin{evalprompt}
You are a strict, consistent evaluator for a full-duplex spoken dialogue benchmark. You will be shown what the model was saying before and during a cue, what it said after the cue ended, and what this condition required. Answer a single yes/no question using ONLY the evidence given. If the evidence is ambiguous or insufficient, answer NO. Return strict JSON only: {"answer": "YES"|"NO", "reason": "<one short sentence>"}.
\end{evalprompt}

\textbf{Condition-specific requirements.}
\begin{evalprompt}
continue_with_background_speech:
Preserve the current content; do not treat background speech as addressed input.

continue_with_other_directed_speech:
Preserve the current content; do not treat other-directed speech as addressed input.

continue_current_response:
Continue the current response without unnecessary change.

continue_but_shorten:
Shorten or condense the ongoing response.

continue_with_simplification:
Simplify the language or content of the ongoing response.
\end{evalprompt}

\textbf{User prompt template.}
\begin{evalprompt}
Action: hold_floor
Condition: {condition_id}

What this condition requires:
{required_adjustment or '(not specified)'}

Cue:
{cue or '(none)'}

Model speech before and during the cue:
{model_through_cue or '(none)'}

Model speech after the cue ended:
{model_after_cue or '(none)'}

Question: Relative to what the model was saying before and during the cue, does its speech after the cue ended meet this condition's requirement? Answer YES if it keeps the same thread and does what the condition asked (keep going, shorten, simplify, or ignore a distractor). Answer NO if it abandons the thread, takes up a cue that should be ignored, or fails to make the change the condition required.

Reply with JSON: {"answer": "YES"|"NO", "reason": "<one short sentence>"}.
\end{evalprompt}
% ============================================================
\subsection{Interrupt}
\label{app:action_interrupt}
% ============================================================

\paragraph{Definition.}
\emph{Interrupt} requires the model to take the floor while the user is still speaking when intervention is needed and to carry out the intervention required by the interaction condition.

\paragraph{Conditions and protocol.}
\emph{Interrupt} is evaluated using Simulator Replay under two conditions: \emph{help} and \emph{safety}. In both conditions, the user utterance is streamed while the model is able to respond concurrently. The same immediate-execution criterion is used for both conditions, while the condition determines the purpose that the subsequent interruption should fulfil.

\paragraph{Immediate execution.}
Immediate execution is successful when the model begins a substantive interruption while the user is still speaking and at least $1.0$\,s before the end of the streamed user utterance. The valid interruption interval is
\[
W_i^{\mathrm{Int}}
=
[t_i^{\mathrm{user,start}},\,
t_i^{\mathrm{user,end}}-1.0\,\mathrm{s}].
\]
Any qualifying substantive onset within this interval constitutes success. An onset within the final $1.0$\,s of the user utterance is treated as overlapping turn entry rather than successful interruption. Speech segments whose complete transcript contains at most three words are treated as backchannels and do not constitute substantive interruptions.

We additionally report \emph{Interruption Timing Difference} as the timing diagnostic for \emph{Interrupt}, as defined in Section~\ref{app:timing_metrics}. The authored reference interruption point is used only for this diagnostic and does not determine immediate-execution success.

\paragraph{Subsequent follow-through.}
Conditional on successful immediate execution, subsequent follow-through evaluates whether the model carries out the intervention required by the condition. For \emph{help}, the interruption should provide the specific assistance or constraint required by the situation. For \emph{safety}, it should identify the relevant safety risk and explain why it is dangerous or what the user should do instead. Merely announcing an interruption, producing a generic response, or failing to address the issue motivating the interruption is unsuccessful. 

\paragraph{Exact evaluation prompt.}

\textbf{System prompt.}
\begin{evalprompt}
You are a strict, consistent evaluator for a full-duplex spoken dialogue benchmark. You will be shown why an interruption is needed, the user's speech, and the model's interruption, and asked a single yes/no question. Answer using ONLY the evidence given. If the evidence is ambiguous or insufficient, answer NO. Return strict JSON only: {"answer": "YES"|"NO", "reason": "<one short sentence>"}.
\end{evalprompt}

\textbf{Condition-specific context.}
\begin{evalprompt}
interrupt_for_safety:
The user is describing a dangerous or hazardous behaviour. An interruption is needed to flag the specific safety risk and to say why it is dangerous or what to do instead.

interrupt_for_help:
The user is proceeding in a direction where timely help would matter. An interruption is needed to offer the specific assistance or constraint the current situation requires.
\end{evalprompt}

\textbf{User prompt template.}
\begin{evalprompt}
Action: interrupt
Condition: {condition_id}

Condition (why an interruption is needed):
{CONDITION_PROMPTS[condition_id] or '(not specified)'}

User speech (what was underway):
{user_speech}

Model interruption:
{model_interruption or '(model produced no speech)'}

Question: Did this interruption do the work this condition required, rather than merely announcing that an interruption is happening? Answer YES if the model's speech addresses the obstacle or action described in the condition. Answer NO if it only announces the interruption, is generic, off-topic, or misses the issue.

Reply with JSON: {"answer": "YES"|"NO", "reason": "<one short sentence>"}.
\end{evalprompt}
% ============================================================
\subsection{Backchannel}
\label{app:action_backchannel}
% ============================================================

\paragraph{Definition.}
\emph{Backchannel} requires the model to provide brief listener feedback of the appropriate function while the user retains the floor, and for its later response to remain consistent with the stance or understanding expressed by that feedback.

\paragraph{Conditions and protocol.}
\emph{Backchannel} is evaluated using Direct Replay under two conditions: \emph{affirmative} and \emph{uncertainty}. \emph{Affirmative} backchannels express acknowledgement, understanding, or agreement, while \emph{uncertainty} backchannels express hesitation, confusion, or a need for clarification. Candidate backchannels are identified during the complete user-speech interval,
\[
W_i^{\mathrm{BC}}
=
[t_i^{\mathrm{user,start}},\,t_i^{\mathrm{user,end}}].
\]
No timing window around the human reference backchannel point is used to determine success.

\paragraph{Immediate execution.}
Immediate execution is determined through three stages. First, we identify the first candidate model segment whose onset falls within the user-speech interval and whose complete transcript contains at most three words. This rule is used only for candidate detection and does not by itself establish successful backchannel production.

Second, a text-based judge determines whether the candidate is a genuine, self-contained backchannel that provides listener feedback while leaving the floor with the user. Candidates that form the truncated beginning of a substantive turn, contain no intelligible speech, or do not express a listener-feedback function are rejected.

Third, for candidates confirmed as genuine backchannels, an audio-capable judge classifies the feedback function as \emph{affirmative} or \emph{uncertain}. The judge does not receive the target condition or the candidate transcript. Immediate execution is successful only when a genuine backchannel is confirmed and its predicted function matches the function specified by the interaction condition.

\paragraph{Subsequent follow-through.}
Conditional on successful immediate execution, subsequent follow-through evaluates whether the model's later substantive speech remains consistent with the understanding or stance expressed by the backchannel. A change in stance is permitted when it is justified by user speech occurring after the backchannel. If no later model speech is produced, subsequent follow-through is unsuccessful.

\paragraph{Exact subsequent-evaluation prompt.}

\textbf{System prompt.}
\begin{evalprompt}
You are a strict, consistent evaluator for a full-duplex spoken dialogue benchmark. You will be shown the user's speech before and overlapping the model's backchannel, the user's speech after that backchannel and before the model's later response, the backchannel itself, and the later model speech. Answer a single yes/no question using ONLY the evidence given. If the evidence is ambiguous or insufficient, answer NO. Return strict JSON only: {"answer": "YES"|"NO", "reason": "<one short sentence>"}.
\end{evalprompt}

\textbf{User prompt template.}
\begin{evalprompt}
Action: backchannel

User speech before and overlapping the model backchannel:
{user_before_and_overlap or '(none)'}

User speech after the backchannel and before the model's later response:
{user_between or '(none)'}

Model backchannel:
{model_backchannel or '(none)'}

Model later speech:
{model_later or '(none)'}

Question: Is the model's later speech consistent with the understanding or stance displayed in its backchannel? Answer YES if later speech fits that backchannel, or is justified by what the user said afterwards. Answer NO if later speech contradicts the backchannel without such justification.

Reply with JSON: {"answer": "YES"|"NO", "reason": "<one short sentence>"}.
\end{evalprompt}
% ============================================================
\subsection{Ignore}
\label{app:action_ignore}
% ============================================================

\paragraph{Definition.}
\emph{Ignore} requires the model to withhold a substantive response to speech that does not require its participation and to prevent that speech from influencing its subsequent response to the addressed user.

\paragraph{Conditions and protocol.}
\emph{Ignore} is evaluated using Direct Replay under two conditions: \emph{background speech} and \emph{other-directed speech}. In both conditions, immediate execution is evaluated over the annotated distractor interval,
\[
W_i^{\mathrm{Ign}}
=
[t_i^{\mathrm{other,start}},\,t_i^{\mathrm{other,end}}].
\]
The same immediate and subsequent criteria are used for both conditions.

\paragraph{Immediate execution.}
Immediate execution is successful when no substantive model response overlaps the distractor interval. A speech segment whose complete transcript contains at most three words is treated as a backchannel and does not constitute a substantive response. 

\paragraph{Subsequent follow-through.}
Conditional on successful immediate execution, subsequent follow-through is successful when the model's later response remains focused on the addressed interaction and does not incorporate the ignored speech. The response should not quote, echo, answer, or otherwise treat the ignored speech as relevant information or a task. This criterion is shared across both conditions and introduces no additional condition-specific constraint.

\paragraph{Exact evaluation prompt.}

\textbf{System prompt.}
\begin{evalprompt}
You are a strict, consistent evaluator for a full-duplex spoken dialogue benchmark. You will be shown the user's addressed speech, other speech that should be ignored, and the model's speech (already restricted to after that other speech ended). Answer a single yes/no question using ONLY the evidence given. If the evidence is ambiguous or insufficient, answer NO. Return strict JSON only: {"answer": "YES"|"NO", "reason": "<one short sentence>"}.
\end{evalprompt}

\textbf{User prompt template.}
\begin{evalprompt}
Action: ignore

User speech addressed to the model:
{addressed_user_speech}

Other speech (should be ignored):
{ignored_speech}

Model speech (after the other speech ended):
{model_response or '(none)'}

Question: Does the model's reply stay free of the other speech? Answer YES if it neither contains that other speech nor is influenced by it: no quoting, echoing, answering, or treating it as a fact or task. Answer NO if the reply includes the other speech or is shaped by it.

Reply with JSON: {"answer": "YES"|"NO", "reason": "<one short sentence>"}.
\end{evalprompt}
% ============================================================
\subsection{Recover}
\label{app:action_recover}
% ============================================================

\paragraph{Definition.}
\emph{Recover} requires the model to re-enter the interaction after an interruption and adapt its resumed contribution to what the interruption requires.

\paragraph{Conditions and protocol.}
\emph{Recover} is evaluated using Simulator Replay under six conditions: \emph{wrap up}, \emph{repeat}, \emph{ask follow-up}, \emph{enrich}, \emph{shift topic}, and \emph{revise}. The interruption is introduced while the model is speaking. All six conditions share the same immediate-execution criterion, while the condition specifies how the resumed contribution should subsequently be adapted.

\paragraph{Immediate execution.}
Immediate execution is successful when the model begins a substantive recovery response within
\[
W_i^{\mathrm{Rec}}
=
[t_i^{\mathrm{interruption,end}}-2.0\,\mathrm{s},\,
t_i^{\mathrm{interruption,end}}+5.0\,\mathrm{s}].
\]
Any qualifying substantive onset within this interval constitutes success and need not be the first model onset observed around the interruption. Speech segments whose complete transcript contains at most three words are treated as backchannels and do not constitute substantive recovery. The negative portion of the interval allows the model to begin re-entering before the interruption has completely ended, while the positive portion permits delayed recovery.

We additionally report \emph{Recovery Latency} as the timing diagnostic for \emph{Recover}, as defined in Section~\ref{app:timing_metrics}.

\paragraph{Subsequent follow-through.}
Conditional on successful immediate execution, subsequent follow-through evaluates whether the recovery makes the adjustment required by the interruption rather than simply resuming the previous contribution unchanged. Depending on the interaction condition, the model should wrap up the ongoing contribution, repeat requested content, ask an appropriate follow-up, enrich the response with requested information, shift to a newly introduced topic, or revise previously stated content.

\paragraph{Exact evaluation prompt.}

\textbf{System prompt.}
\begin{evalprompt}
You are a strict, consistent evaluator for a full-duplex spoken dialogue benchmark. You will be shown the interrupted contribution, the user's interruption, and the model's recovery in the observation window, and asked a single yes/no question. Answer using ONLY the evidence given. If the evidence is ambiguous or insufficient, answer NO. Return strict JSON only: {"answer": "YES"|"NO", "reason": "<one short sentence>"}.
\end{evalprompt}

\textbf{User prompt template.}
\begin{evalprompt}
Action: recover

Interrupted contribution (what the model was saying):
{interrupted_contribution or '(not shown)'}

User interruption:
{interruption}

Model recovery (speech with onset in the observation window):
{model_recovery or '(model produced no speech)'}

Question: Does the model's response in the observation window make the adjustment this interruption required? Answer YES if it adapts as the user's interruption called for, given what the model was saying. Answer NO if it ignores the interruption, adjusts incorrectly, or resumes obsolete content as if the interruption had not happened.

Reply with JSON: {"answer": "YES"|"NO", "reason": "<one short sentence>"}.
\end{evalprompt}
\section{Interaction Pipeline Details}
\label{app:interaction_pipeline}

This section provides additional details of the two interaction pipelines introduced in Section~\ref{sec:interaction_pipeline}. 

\subsection{Direct Replay}
\label{app:direct_replay}

Each Direct Replay instance is evaluated in an independent session using a fixed user-audio input. The input is streamed in real time according to the model's native input configuration, while model output is recorded concurrently on the same time-aligned input--output timeline. The timing origin is defined by the first streamed input sample rather than connection establishment, ensuring that interaction timestamps are comparable across model interfaces. A fixed 15\,s observation period is appended after the user content to capture responses following the target interaction point.

Model speech segments used for evaluation are derived from the recorded output timeline rather than directly from provider-side turn-detection events. Provider-side voice activity or turn detection may control when a model begins generating a response, but does not define the scored response onset. This allows the action-specific timing criteria in Appendix~\ref{app:action_evaluation} to be applied consistently across different model interfaces.

\subsection{Simulator-Based Interaction}
\label{app:simulator_interaction}

Simulator-based evaluation retains fixed benchmark audio while adapting interaction timing to the model's ongoing behaviour. The simulator uses GPT-5.4 as a decision model and operates only as an interaction controller; it does not generate the content or speech of the simulated user online. All user cues are prerecorded benchmark audio, allowing their linguistic and acoustic content to remain fixed across evaluated models.

\paragraph{Cue injection for Stop, Hold Floor, and Recover.}
These actions use a lead-in followed by a prerecorded interaction cue. The lead-in is first streamed to elicit an ongoing model response. Once the model is speaking, the simulator observes its model-provided streaming transcript together with the reference response and cue information associated with the benchmark instance, and determines an appropriate point at which to inject the fixed cue. If no suitable point is identified within the permitted interaction period, the cue is force-injected to ensure that the benchmark opportunity is presented. A force-injected trial is not automatically treated as invalid; its inclusion in action evaluation depends on whether the cue actually occurs while the model is speaking.

For model interfaces whose streaming transcript can advance ahead of audible output, the transcript exposed to the simulator is aligned with the progression of model playback. This prevents the simulator from making an insertion decision using model content that has not yet reached the corresponding interaction point, while avoiding an additional speech-recognition stage over the rendered audio. After cue playback, the interaction continues for a fixed observation period to capture the model's immediate and subsequent behaviour.

\paragraph{Barge-in interaction for Interrupt.}
For \emph{Interrupt}, the interaction direction is reversed: the simulated user continuously holds the floor while the model is allowed to intervene. The simulator monitors the model's ongoing output and determines whether detected model speech represents an attempt to take the turn rather than a brief backchannel or filler. When a takeover is detected, the remaining unsent user audio is stopped so that the interaction can proceed from the model's intervention; otherwise, the user utterance continues to completion.

The simulator decision in this pipeline controls only the progression of the interaction and is not itself used as the Interrupt evaluation outcome. Immediate execution is independently computed from the recorded model-speech timeline using the timing criterion defined in Appendix~\ref{app:action_interrupt}.

\section{Model Setup}
\label{app:model_setup}

All evaluated models use their officially released implementations and recommended inference configurations. For open-source models, we use the publicly released checkpoints listed in Table~\ref{tab:model_versions}; for close-source models, we report the API model identifiers used during evaluation. Our evaluation interfaces adapt these implementations only as required to connect them to the common interaction pipelines described in Section~\ref{sec:interaction_pipeline}.

\begin{table*}[t]
\centering
\caption{Model versions and checkpoints used in our experiments.}
\label{tab:model_versions}
\small
\setlength{\tabcolsep}{5pt}
\renewcommand{\arraystretch}{1.10}

\begin{tabular}{lll}
\toprule
\textbf{Model}
&
\textbf{Access}
&
\textbf{Version / Checkpoint}
\\
\midrule

Moshi
&
Open-source
&
\texttt{kyutai/moshiko-pytorch-bf16}
\\

Freeze-Omni
&
Open-source
&
\texttt{VITA-MLLM/Freeze-Omni}
\\

Mini-Omni2
&
Open-source
&
\texttt{gpt-omni/mini-omni2}
\\

VITA-1.5
&
Open-source
&
\texttt{VITA-MLLM/VITA-1.5}
\\

Voila-Autonomous
&
Open-source
&
\texttt{maitrix-org/Voila-autonomous-preview}
\\

FLM-Audio
&
Open-source
&
\texttt{CofeAI/FLM-Audio}
\\

Raon-SpeechChat
&
Open-source
&
\texttt{KRAFTON/Raon-SpeechChat-9B}
\\

BayLing-Duplex
&
Open-source
&
\texttt{BayLing-Models/BayLing-Duplex}
\\

Lychee-FD
&
Open-source
&
\texttt{HIT-TMG/Lychee-FD}
\\

MiniCPM-o 4.5
&
Open-source
&
\texttt{openbmb/MiniCPM-o-4\_5}
\\

\midrule

Gemini 3.1 Flash Live
&
Close-source
&
\texttt{gemini-3.1-flash-live-preview}
\\

GPT-Realtime-2.1
&
Close-source
&
\texttt{gpt-realtime-2.1}
\\

Grok Voice
&
Close-source
&
\texttt{grok-voice-latest}\footnotemark
\\

Qwen3.5-Omni-Flash-Realtime
&
Close-source
&
\texttt{qwen3.5-omni-flash-realtime}
\\

\midrule
\multicolumn{3}{l}{\textit{Controlled synthesis}} \\

GPT-5.4
&
Close-source
&
\texttt{gpt-5.4-2026-03-05}
\\

ElevenLabs v3
&
Close-source
&
\texttt{eleven\_v3}
\\

\bottomrule
\end{tabular}
\end{table*}

\footnotetext{\texttt{grok-voice-latest} is a floating API model identifier. Close-source evaluations were conducted using the model versions served by their respective APIs during September 6--13, 2026.}

\subsection{Model Output Transcription}
\label{app:model_transcription}

For transcript-dependent evaluation, we use \texttt{whisper-1} to transcribe recorded model speech for the majority of evaluated models. For \emph{BayLing-Duplex} and \emph{Raon-SpeechChat}, we instead use their native streaming transcripts because temporal jitter in the rendered audio makes post-hoc ASR transcription less reliable for transcript-dependent evaluation.
% Condition-level I / S results
\newcommand{\condscr}[3]{#1/#2\, (#3)}
\newcommand{\condna}[1]{#1/\texttt{n/a}}
% Timing diagnostics: median (IQR, n)
\newcommand{\timingcell}[3]{#1\, (#2, #3)}
\newcommand{\timingsingle}[2]{#1\, (\texttt{n/a}, #2)}
\newcommand{\timingna}{\texttt{n/a}}

% ============================================================
\section{Comprehensive Condition-Level Results}
\label{app:comprehensive_results}
% ============================================================

Table~\ref{tab:full_condition_results} reports the complete condition-level results for all evaluated models. Each cell reports \emph{immediate execution} / \emph{subsequent follow-through}, with the number of instances entering subsequent evaluation shown in parentheses. Subsequent follow-through is evaluated conditional on successful immediate execution and any action-specific applicability requirements; \texttt{n/a} indicates that no instance entered the corresponding subsequent evaluation. Table~\ref{tab:full_timing_results} separately reports the timing diagnostics for \emph{Stop}, \emph{Recover}, and \emph{Interrupt} as median (IQR, $n_T$) in seconds.

\begin{table*}[t]
\centering
\caption{
Condition-level timing diagnostics in seconds. Each cell reports median
(IQR, $n_T$). For \emph{Stop}, smaller values indicate faster yielding.
For \emph{Recover}, values are relative to interruption end.
For \emph{Interrupt}, negative values indicate intervention before the
authored target point.
}
\label{tab:full_timing_results}

\scriptsize
\renewcommand{\arraystretch}{1.0}

% ============================================================
% (a) Stop
% ============================================================

\begin{minipage}[t]{0.38\textwidth}
\centering

\textbf{(a) Stop: Yield Latency}

\vspace{2pt}

\setlength{\tabcolsep}{2pt}
\begin{tabular*}{\linewidth}{@{\extracolsep{\fill}}lc@{}}
\toprule
\textbf{Model}
& \textbf{Direct interruption}
\\
\midrule

Moshi
& \timingcell{0.96}{1.97}{96}
\\
Freeze-Omni
& \timingcell{1.80}{3.37}{90}
\\
FLM-Audio
& \timingcell{4.13}{5.46}{146}
\\
Raon-SpeechChat
& \timingcell{5.20}{5.06}{116}
\\
BayLing-Duplex
& \timingcell{3.05}{1.55}{161}
\\
MiniCPM-o 4.5
& \timingcell{3.47}{2.74}{167}
\\

\midrule

Gemini 3.1 Flash Live
& \timingcell{1.74}{0.31}{195}
\\
GPT-Realtime-2.1
& \timingcell{1.61}{0.22}{195}
\\
Grok Voice
& \timingcell{2.37}{0.72}{171}
\\
Qwen3.5-Omni-Flash-Realtime
& \timingcell{2.41}{0.82}{184}
\\

\bottomrule
\end{tabular*}

\end{minipage}
\hfill
% ============================================================
% (b) Interrupt
% ============================================================
\begin{minipage}[t]{0.59\textwidth}
\centering

\textbf{(b) Interrupt: Interruption Timing Difference}

\vspace{2pt}

\setlength{\tabcolsep}{2pt}
\begin{tabular*}{\linewidth}{@{\extracolsep{\fill}}lcc@{}}
\toprule
\textbf{Model}
& \textbf{Help}
& \textbf{Safety}
\\
\midrule

Moshi
& \timingcell{-7.53}{11.14}{178}
& \timingcell{-0.28}{4.11}{92}
\\
Freeze-Omni
& \timingcell{-10.54}{8.71}{81}
& \timingcell{-0.95}{3.11}{21}
\\
FLM-Audio
& \timingcell{-6.41}{6.65}{9}
& \timingcell{0.65}{2.40}{2}
\\
Raon-SpeechChat
& \timingcell{-16.52}{10.71}{75}
& \timingcell{-5.63}{2.26}{72}
\\
BayLing-Duplex
& \timingcell{-5.72}{8.15}{86}
& \timingcell{-1.50}{2.15}{31}
\\
MiniCPM-o 4.5
& \timingcell{-7.33}{6.79}{25}
& \timingcell{-0.35}{2.67}{14}
\\

\midrule

Gemini 3.1 Flash Live
& \timingcell{-5.48}{6.13}{7}
& \timingna
\\
GPT-Realtime-2.1
& \timingcell{-11.23}{10.25}{23}
& \timingcell{-1.93}{3.20}{12}
\\
Grok Voice
& \timingna
& \timingna
\\
Qwen3.5-Omni-Flash-Realtime
& \timingcell{-6.19}{9.21}{8}
& \timingsingle{0.09}{1}
\\

\bottomrule
\end{tabular*}

\end{minipage}

\vspace{7pt}

% ============================================================
% (c) Recover
% ============================================================

\textbf{(c) Recover: Recovery Latency}

\vspace{2pt}

\setlength{\tabcolsep}{3pt}
\begin{tabular*}{\textwidth}{@{\extracolsep{\fill}}lcccccc@{}}
\toprule
\textbf{Model}
& \textbf{Wrap up}
& \textbf{Repeat}
& \textbf{Ask follow-up}
& \textbf{Enrich}
& \textbf{Shift topic}
& \textbf{Revise}
\\
\midrule

Moshi
& \timingcell{0.86}{2.58}{77}
& \timingcell{1.30}{2.36}{96}
& \timingcell{1.24}{2.62}{98}
& \timingcell{1.41}{2.20}{71}
& \timingcell{1.49}{1.43}{67}
& \timingcell{1.47}{3.23}{55}
\\

Freeze-Omni
& \timingcell{-1.83}{0.14}{2}
& \timingcell{0.64}{2.40}{2}
& \timingcell{2.51}{2.42}{3}
& \timingcell{2.52}{1.02}{9}
& \timingcell{2.15}{0.53}{13}
& \timingcell{1.76}{0.70}{8}
\\

FLM-Audio
& \timingna
& \timingsingle{-1.98}{1}
& \timingsingle{-1.91}{1}
& \timingcell{4.02}{3.99}{9}
& \timingcell{4.52}{0.95}{12}
& \timingcell{4.67}{0.54}{12}
\\

Raon-SpeechChat
& \timingcell{2.21}{3.68}{10}
& \timingcell{0.99}{0.46}{3}
& \timingcell{0.40}{1.92}{5}
& \timingcell{2.33}{4.53}{28}
& \timingcell{2.70}{2.84}{18}
& \timingcell{3.83}{1.93}{19}
\\

BayLing-Duplex
& \timingcell{3.69}{1.64}{62}
& \timingcell{3.66}{1.17}{69}
& \timingcell{3.64}{1.48}{50}
& \timingcell{3.15}{1.80}{64}
& \timingcell{3.50}{2.19}{60}
& \timingcell{3.68}{1.82}{75}
\\

MiniCPM-o 4.5
& \timingcell{3.10}{1.08}{37}
& \timingcell{2.58}{1.07}{66}
& \timingcell{2.43}{1.03}{75}
& \timingcell{2.39}{1.24}{106}
& \timingcell{2.41}{0.91}{109}
& \timingcell{2.42}{1.22}{102}
\\

\midrule

Gemini 3.1 Flash Live
& \timingcell{2.53}{0.26}{188}
& \timingcell{2.58}{0.32}{194}
& \timingcell{2.55}{0.24}{193}
& \timingcell{2.59}{0.35}{192}
& \timingcell{2.59}{0.34}{197}
& \timingcell{2.54}{0.30}{185}
\\

GPT-Realtime-2.1
& \timingcell{2.84}{0.53}{151}
& \timingcell{2.93}{0.47}{148}
& \timingcell{2.98}{0.46}{149}
& \timingcell{3.08}{0.56}{141}
& \timingcell{3.05}{0.72}{149}
& \timingcell{3.11}{0.57}{152}
\\

Grok Voice
& \timingcell{3.02}{0.47}{164}
& \timingcell{3.11}{0.41}{174}
& \timingcell{3.04}{0.41}{161}
& \timingcell{3.20}{0.42}{131}
& \timingcell{3.22}{0.37}{152}
& \timingcell{3.17}{0.48}{118}
\\

Qwen3.5-Omni-Flash-Realtime
& \timingcell{2.68}{0.37}{174}
& \timingcell{2.66}{0.33}{180}
& \timingcell{2.65}{0.29}{175}
& \timingcell{2.69}{0.36}{164}
& \timingcell{2.68}{0.41}{168}
& \timingcell{2.71}{0.41}{150}
\\

\bottomrule
\end{tabular*}

\end{table*}

\begin{table*}[t]
\centering
\caption{
Complete condition-level results. Each cell reports I / S ($n_S$), in percentage, where $n_S$ is the number of instances entering subsequent follow-through evaluation.
}
\label{tab:full_condition_results}

\scriptsize
\setlength{\tabcolsep}{1.4pt}
\renewcommand{\arraystretch}{0.94}

% ============================================================
% (a) Direct Replay: Wait + Ignore
% ============================================================

\begin{tabular*}{\textwidth}{@{\extracolsep{\fill}}lccccc@{}}
\toprule
\multicolumn{6}{c}{\textbf{(a) Direct Replay: Wait and Ignore}} \\
\midrule
&
\multicolumn{3}{c}{\textbf{Wait}}
&
\multicolumn{2}{c}{\textbf{Ignore}}
\\
\cmidrule(lr){2-4}
\cmidrule(lr){5-6}

\textbf{Model}
& \textbf{Inter-sentential}
& \textbf{Intra-sentential}
& \textbf{Continuous}
& \textbf{Background}
& \textbf{Other-directed}
\\
\midrule

Moshi
& \condscr{81.5}{42.3}{163}
& \condscr{90.0}{47.8}{180}
& \condscr{75.0}{46.7}{150}
& \condscr{43.0}{89.5}{86}
& \condscr{22.0}{88.6}{44}
\\
Freeze-Omni
& \condscr{67.0}{14.9}{134}
& \condscr{83.5}{16.8}{167}
& \condscr{69.0}{39.1}{138}
& \condscr{58.5}{8.5}{117}
& \condscr{64.5}{10.9}{129}
\\
Mini-Omni2
& \condscr{10.5}{19.0}{21}
& \condscr{13.5}{33.3}{27}
& \condscr{2.0}{100.0}{4}
& \condscr{2.0}{0.0}{4}
& \condscr{21.5}{2.3}{43}
\\
VITA-1.5
& \condscr{95.5}{56.0}{191}
& \condscr{93.5}{59.9}{187}
& \condscr{86.5}{78.6}{173}
& \condscr{65.0}{100.0}{130}
& \condscr{46.0}{79.3}{92}
\\
Voila-Autonomous
& \condscr{95.0}{34.7}{190}
& \condscr{98.0}{45.4}{196}
& \condscr{95.5}{57.1}{191}
& \condscr{98.5}{98.5}{197}
& \condscr{99.5}{66.8}{199}
\\
FLM-Audio
& \condscr{94.5}{14.8}{189}
& \condscr{98.5}{19.3}{197}
& \condscr{93.5}{20.9}{187}
& \condscr{99.0}{87.4}{198}
& \condscr{69.5}{87.8}{139}
\\
Raon-SpeechChat
& \condscr{46.5}{63.4}{93}
& \condscr{65.0}{71.5}{130}
& \condscr{43.5}{75.9}{87}
& \condscr{67.5}{94.8}{135}
& \condscr{26.5}{41.5}{53}
\\
BayLing-Duplex
& \condscr{90.5}{26.5}{181}
& \condscr{96.0}{33.9}{192}
& \condscr{92.0}{33.2}{184}
& \condscr{82.0}{92.7}{164}
& \condscr{74.0}{89.9}{148}
\\
Lychee-FD
& \condscr{98.0}{51.0}{196}
& \condscr{99.5}{54.3}{199}
& \condscr{97.0}{54.1}{194}
& \condscr{93.5}{100.0}{187}
& \condscr{68.0}{94.1}{136}
\\
MiniCPM-o 4.5
& \condscr{94.0}{68.1}{188}
& \condscr{98.5}{73.6}{197}
& \condscr{93.0}{81.7}{186}
& \condscr{86.5}{100.0}{173}
& \condscr{59.0}{60.2}{118}
\\
\midrule
Gemini 3.1 Flash Live
& \condscr{98.0}{79.1}{196}
& \condscr{100.0}{85.5}{200}
& \condscr{96.0}{91.7}{192}
& \condscr{94.0}{95.2}{188}
& \condscr{64.0}{60.9}{128}
\\
GPT-Realtime-2.1
& \condscr{100.0}{91.5}{200}
& \condscr{100.0}{96.0}{200}
& \condscr{99.5}{99.5}{199}
& \condscr{92.0}{94.6}{184}
& \condscr{95.0}{53.7}{190}
\\
Grok Voice
& \condscr{99.0}{72.2}{198}
& \condscr{100.0}{78.0}{200}
& \condscr{98.5}{72.6}{197}
& \condscr{99.5}{99.5}{199}
& \condscr{97.5}{50.8}{195}
\\
Qwen3.5-Omni-Flash-Realtime
& \condscr{86.5}{85.0}{173}
& \condscr{95.5}{92.1}{191}
& \condscr{86.0}{94.2}{172}
& \condscr{89.0}{100.0}{178}
& \condscr{48.0}{42.7}{96}
\\
\bottomrule
\end{tabular*}

\vspace{1pt}

% ============================================================
% (b) Direct Replay: Take Turn + Backchannel
% ============================================================

\begin{tabular*}{\textwidth}{@{\extracolsep{\fill}}lccccc@{}}
\toprule
\multicolumn{6}{c}{\textbf{(b) Direct Replay: Take Turn and Backchannel}} \\
\midrule
&
\multicolumn{3}{c}{\textbf{Take Turn}}
&
\multicolumn{2}{c}{\textbf{Backchannel}}
\\
\cmidrule(lr){2-4}
\cmidrule(lr){5-6}

\textbf{Model}
& \textbf{Immediate}
& \textbf{Delayed}
& \textbf{Slight overlap}
& \textbf{Affirmative}
& \textbf{Uncertainty}
\\
\midrule

Moshi
& \condscr{33.5}{19.4}{67}
& \condscr{53.0}{10.4}{106}
& \condscr{22.0}{40.9}{44}
& \condscr{10.5}{66.7}{21}
& \condscr{2.0}{25.0}{4}
\\
Freeze-Omni
& \condscr{2.0}{50.0}{4}
& \condscr{5.0}{30.0}{10}
& \condscr{23.5}{51.1}{47}
& \condscr{0.5}{0.0}{1}
& \condna{0.0}
\\
Mini-Omni2
& \condscr{1.0}{0.0}{2}
& \condscr{12.0}{0.0}{24}
& \condscr{23.0}{23.9}{46}
& \condna{0.0}
& \condna{0.0}
\\
VITA-1.5
& \condscr{2.5}{0.0}{5}
& \condscr{17.0}{26.5}{34}
& \condscr{4.5}{44.4}{9}
& \condna{0.0}
& \condna{0.0}
\\
Voila-Autonomous
& \condna{0.0}
& \condscr{98.0}{20.4}{196}
& \condscr{35.0}{20.0}{70}
& \condscr{3.0}{33.3}{6}
& \condna{0.0}
\\
FLM-Audio
& \condscr{0.5}{0.0}{1}
& \condscr{86.5}{1.7}{173}
& \condscr{18.5}{13.5}{37}
& \condscr{0.5}{0.0}{1}
& \condna{0.0}
\\
Raon-SpeechChat
& \condscr{63.5}{30.7}{127}
& \condscr{69.5}{14.4}{139}
& \condscr{62.5}{24.8}{125}
& \condscr{43.0}{46.5}{86}
& \condscr{19.0}{36.8}{38}
\\
BayLing-Duplex
& \condscr{3.5}{14.3}{7}
& \condscr{12.0}{0.0}{24}
& \condscr{19.5}{15.4}{39}
& \condscr{0.5}{0.0}{1}
& \condna{0.0}
\\
Lychee-FD
& \condscr{0.5}{0.0}{1}
& \condscr{59.0}{11.0}{118}
& \condscr{3.0}{0.0}{6}
& \condscr{2.5}{20.0}{5}
& \condscr{2.0}{25.0}{4}
\\
MiniCPM-o 4.5
& \condscr{4.5}{22.2}{9}
& \condscr{85.5}{41.5}{171}
& \condscr{23.5}{53.2}{47}
& \condscr{4.0}{62.5}{8}
& \condscr{1.0}{100.0}{2}
\\
\midrule
Gemini 3.1 Flash Live
& \condna{0.0}
& \condscr{96.5}{79.8}{193}
& \condscr{18.5}{94.6}{37}
& \condscr{2.0}{50.0}{4}
& \condna{0.0}
\\
GPT-Realtime-2.1
& \condna{0.0}
& \condscr{76.0}{84.2}{152}
& \condscr{6.5}{69.2}{13}
& \condscr{0.5}{100.0}{1}
& \condna{0.0}
\\
Grok Voice
& \condscr{0.5}{100.0}{1}
& \condscr{84.5}{80.5}{169}
& \condscr{6.0}{100.0}{12}
& \condscr{1.5}{33.3}{3}
& \condna{0.0}
\\
Qwen3.5-Omni-Flash-Realtime
& \condna{0.0}
& \condscr{97.5}{75.4}{195}
& \condscr{26.5}{90.6}{53}
& \condscr{6.5}{92.3}{13}
& \condscr{1.0}{50.0}{2}
\\
\bottomrule
\end{tabular*}

\vspace{1pt}

% ============================================================
% (c) Simulator Replay: Stop + Hold Floor + Interrupt Help
% ============================================================

{\fontsize{6.5pt}{6.8pt}\selectfont

\begin{tabular*}{\textwidth}{@{\extracolsep{\fill}}lccccccc@{}}
\toprule
\multicolumn{8}{c}{\textbf{(c) Simulator Replay: Stop, Hold Floor, and Interrupt}} \\
\midrule
&
\multicolumn{1}{c}{\textbf{Stop}}
&
\multicolumn{5}{c}{\textbf{Hold Floor}}
&
\multicolumn{1}{c}{\textbf{Interrupt}}
\\
\cmidrule(lr){2-2}
\cmidrule(lr){3-7}
\cmidrule(lr){8-8}

\textbf{Model}
& \textbf{Direct}
& \textbf{Continue}
& \textbf{Shorten}
& \textbf{Simplify}
& \textbf{Background}
& \textbf{Other-dir.}
& \textbf{Help}
\\
\midrule

Moshi
& \condscr{98.0}{70.7}{82}
& \condscr{65.1}{84.1}{69}
& \condscr{43.9}{8.3}{36}
& \condscr{51.9}{7.3}{55}
& \condscr{39.2}{88.7}{62}
& \condscr{37.8}{77.8}{45}
& \condscr{89.0}{0.0}{178}
\\

Freeze-Omni
& \condscr{94.7}{100.0}{57}
& \condscr{91.3}{70.2}{94}
& \condscr{89.7}{0.0}{52}
& \condscr{93.6}{1.0}{102}
& \condscr{70.0}{100.0}{63}
& \condscr{64.6}{98.1}{53}
& \condscr{40.5}{23.5}{81}
\\

FLM-Audio
& \condscr{97.3}{100.0}{60}
& \condscr{91.2}{73.1}{93}
& \condscr{88.3}{0.0}{53}
& \condscr{88.9}{0.0}{64}
& \condscr{63.5}{99.1}{106}
& \condscr{67.4}{100.0}{97}
& \condscr{4.5}{0.0}{9}
\\

Raon-SpeechChat
& \condscr{95.1}{93.1}{29}
& \condscr{95.5}{53.1}{128}
& \condscr{93.9}{2.6}{77}
& \condscr{98.4}{2.4}{127}
& \condscr{79.4}{44.4}{54}
& \condscr{76.2}{37.5}{32}
& \condscr{37.5}{0.0}{75}
\\

BayLing-Duplex
& \condscr{98.8}{93.2}{74}
& \condscr{96.6}{51.4}{144}
& \condscr{93.7}{1.0}{104}
& \condscr{96.5}{5.4}{167}
& \condscr{53.1}{65.4}{52}
& \condscr{44.3}{45.2}{31}
& \condscr{43.0}{0.0}{86}
\\

MiniCPM-o 4.5
& \condscr{98.2}{98.8}{86}
& \condscr{91.5}{59.0}{161}
& \condscr{87.3}{0.0}{96}
& \condscr{87.3}{1.4}{145}
& \condscr{82.8}{89.6}{164}
& \condscr{80.6}{79.2}{125}
& \condscr{12.5}{20.0}{25}
\\

\midrule

Gemini 3.1 Flash Live
& \condscr{100.0}{100.0}{150}
& \condscr{93.9}{91.8}{170}
& \condscr{92.4}{11.3}{97}
& \condscr{95.7}{10.2}{157}
& \condscr{46.0}{100.0}{92}
& \condscr{31.9}{98.0}{51}
& \condscr{3.5}{14.3}{7}
\\

GPT-Realtime-2.1
& \condscr{100.0}{100.0}{156}
& \condscr{93.8}{80.3}{183}
& \condscr{89.2}{0.0}{99}
& \condscr{95.1}{12.1}{173}
& \condscr{9.4}{88.9}{18}
& \condscr{3.2}{80.0}{5}
& \condscr{11.5}{13.0}{23}
\\

Grok Voice
& \condscr{100.0}{96.7}{123}
& \condscr{92.3}{76.9}{108}
& \condscr{83.8}{19.4}{62}
& \condscr{89.1}{14.6}{82}
& \condscr{63.8}{63.0}{127}
& \condscr{35.0}{50.0}{56}
& \condna{0.0}
\\

Qwen3.5-Omni-Flash-Realtime
& \condscr{100.0}{98.3}{119}
& \condscr{93.4}{93.6}{156}
& \condscr{92.0}{2.5}{81}
& \condscr{92.7}{2.4}{127}
& \condscr{38.1}{66.7}{69}
& \condscr{60.4}{70.1}{87}
& \condscr{4.0}{75.0}{8}
\\

\bottomrule
\end{tabular*}

\vspace{1pt}

% ============================================================
% (d) Simulator Replay: Recover + Interrupt Safety
% ============================================================

\begin{tabular*}{\textwidth}{@{\extracolsep{\fill}}lccccccc@{}}
\toprule
\multicolumn{8}{c}{\textbf{(d) Simulator Replay: Recover and Interrupt}} \\
\midrule
&
\multicolumn{6}{c}{\textbf{Recover}}
&
\multicolumn{1}{c}{\textbf{Interrupt}}
\\
\cmidrule(lr){2-7}
\cmidrule(lr){8-8}

\textbf{Model}
& \textbf{Wrap up}
& \textbf{Repeat}
& \textbf{Ask follow-up}
& \textbf{Enrich}
& \textbf{Shift topic}
& \textbf{Revise}
& \textbf{Safety}
\\
\midrule

Moshi
& \condscr{55.4}{31.2}{77}
& \condscr{62.3}{17.7}{96}
& \condscr{67.1}{21.4}{98}
& \condscr{64.5}{31.0}{71}
& \condscr{66.3}{52.2}{67}
& \condscr{50.5}{20.0}{55}
& \condscr{46.0}{0.0}{92}
\\

Freeze-Omni
& \condscr{1.7}{0.0}{2}
& \condscr{1.7}{0.0}{2}
& \condscr{2.9}{33.3}{3}
& \condscr{8.6}{66.7}{9}
& \condscr{13.4}{53.8}{13}
& \condscr{7.8}{50.0}{8}
& \condscr{10.5}{4.8}{21}
\\

FLM-Audio
& \condna{0.0}
& \condscr{0.8}{0.0}{1}
& \condscr{0.7}{0.0}{1}
& \condscr{6.0}{22.2}{9}
& \condscr{7.5}{25.0}{12}
& \condscr{8.7}{0.0}{12}
& \condscr{1.0}{0.0}{2}
\\

Raon-SpeechChat
& \condscr{13.9}{10.0}{10}
& \condscr{5.0}{0.0}{3}
& \condscr{8.2}{0.0}{5}
& \condscr{22.2}{7.1}{28}
& \condscr{13.8}{5.6}{18}
& \condscr{14.7}{10.5}{19}
& \condscr{36.0}{0.0}{72}
\\

BayLing-Duplex
& \condscr{54.9}{11.3}{62}
& \condscr{63.3}{11.6}{69}
& \condscr{65.8}{26.0}{50}
& \condscr{43.0}{17.2}{64}
& \condscr{35.7}{26.7}{60}
& \condscr{46.6}{10.7}{75}
& \condscr{15.5}{0.0}{31}
\\

MiniCPM-o 4.5
& \condscr{18.6}{59.5}{37}
& \condscr{33.3}{54.5}{66}
& \condscr{38.1}{76.0}{75}
& \condscr{62.0}{69.8}{106}
& \condscr{63.7}{79.8}{109}
& \condscr{57.3}{53.9}{102}
& \condscr{7.0}{14.3}{14}
\\

\midrule

Gemini 3.1 Flash Live
& \condscr{94.0}{52.1}{188}
& \condscr{97.0}{66.5}{194}
& \condscr{97.0}{88.6}{193}
& \condscr{98.0}{85.9}{192}
& \condscr{98.5}{94.9}{197}
& \condscr{97.9}{82.7}{185}
& \condna{0.0}
\\

GPT-Realtime-2.1
& \condscr{81.6}{79.5}{151}
& \condscr{80.0}{83.1}{148}
& \condscr{78.8}{96.0}{149}
& \condscr{72.3}{95.0}{141}
& \condscr{75.6}{98.7}{149}
& \condscr{78.4}{90.1}{152}
& \condscr{6.0}{0.0}{12}
\\

Grok Voice
& \condscr{82.0}{80.5}{164}
& \condscr{87.0}{59.8}{174}
& \condscr{80.9}{89.4}{161}
& \condscr{77.1}{93.1}{131}
& \condscr{84.0}{97.4}{152}
& \condscr{70.2}{89.8}{118}
& \condna{0.0}
\\

Qwen3.5-Omni-Flash-Realtime
& \condscr{91.1}{82.2}{174}
& \condscr{94.2}{61.7}{180}
& \condscr{92.1}{94.3}{175}
& \condscr{92.1}{93.9}{164}
& \condscr{92.3}{95.8}{168}
& \condscr{85.7}{82.7}{150}
& \condscr{0.5}{0.0}{1}
\\

\bottomrule
\end{tabular*}

} % end smaller font for panels (c) and (d)

\end{table*}

\end{document}